\documentclass{article} % For LaTeX2e
\usepackage{colm2024_conference}
\colmfinalcopy

\usepackage{microtype}
\usepackage{hyperref}
\usepackage{url}

\usepackage{bm}

\usepackage{multicol}
\usepackage{multirow}
\usepackage{booktabs}

\usepackage{xspace}
\usepackage{graphicx}
\usepackage{pgfplots}

\newcommand{\tif}[1]{\textbf{\textit{#1}}}
\usepackage{adjustbox}
\usepackage{microtype}
\usepackage{verbatim}
\usepackage{multirow}
\usepackage{threeparttable}
\usepackage{booktabs}
\usepackage{subfig}
\usepackage{comment}
\usepackage{colortbl} % 提供 \rowcolor
\usepackage{xcolor}   % 提供色彩支持
\usepackage{amsmath}
\usepackage{xspace}
\usepackage[figuresright]{rotating}
\usepackage{algorithm}
\usepackage{algorithmic}
\usepackage{wrapfig}        
\definecolor{arrowred}{rgb}{0.9, 0, 0}    % RGB(203, 0, 0)
\definecolor{arrowgreen}{rgb}{0, 0.9, 0.7}  % RGB(96, 218, 168)

\newcommand{\ourdataset}{\textsc{MbtiBench}\xspace}

\title{Can Large Language Models Understand You Better?\\  An MBTI Personality Detection Dataset Aligned with Population Traits}

\author{
Bohan Li$^{1}$, Jiannan Guan$^{1}$, Longxu Dou$^{2}$, Yunlong Feng$^{1}$, Dingzirui Wang$^{1}$,\\ 
\textbf{ Yang Xu$^{1}$, Enbo Wang$^{1}$, Qiguang Chen$^{1}$, Bichen Wang$^{1}$, Xiao Xu$^{1}$, Yimeng Zhang$^{2}$,} \\
\textbf{ Libo Qin$^{3}$, Yanyan Zhao$^{1}$, Qingfu Zhu$^{1}$, Wanxiang Che$^{1}$\thanks{The corresponding author.}}\\
$^{1}$ Harbin Institute of Technology  \\
$^{2}$ Individual Researcher \\
$^{3}$ Central South University \\
\texttt{\{bhli, jnguan\}@ir.hit.edu.cn}, \texttt{car@ir.hit.edu.cn}* \\\\
}

\begin{document}
    \maketitle
    
\begin{abstract}

The Myers-Briggs Type Indicator (MBTI) is one of the most influential personality theories reflecting individual differences in thinking, feeling, and behaving.
MBTI personality detection has garnered considerable research interest and has evolved significantly over the years.
However, this task tends to be overly optimistic, as it currently does not align well with the natural distribution of population personality traits.
Specifically, (1) the self-reported labels in existing datasets result in incorrect labeling issues,  and (2) the hard labels fail to capture the full range of population personality distributions.
In this paper, we optimize the task by constructing ~\ourdataset, the first manually annotated high-quality MBTI personality detection dataset with soft labels, under the guidance of psychologists.
As for the first challenge, \ourdataset effectively solves the incorrect labeling issues, which account for 29.58\% of the data.
As for the second challenge, we estimate soft labels by deriving the polarity tendency of samples.
The obtained soft labels confirm that there are more people with non-extreme personality traits.
Experimental results not only highlight the polarized predictions and biases in LLMs as key directions for future research, but also confirm that soft labels can provide more benefits to other psychological tasks than hard labels.\footnote{\url{https://github.com/Personality-NLP/MbtiBench}}

\end{abstract}

\section{Introduction}%\footnotetext{Footnotetext without footnote mark}
    Personality, a key psychological concept, refers to individual differences in thinking, feeling, and behavior~\citep{corr2009cambridge}.
Among personality models, the Myers-Briggs Type Indicator (MBTI) is one of the most recognized non-clinical frameworks, with broad applications in areas like stance detection~\citep{Hosseinia2021OnTU}.
Recently, automatically detecting a person’s MBTI type from their written content (e.g., tweets and blogs)~\citep{Plank2015PersonalityTO, pandora} has become a growing area of research with both academic significance~\citep{stajner-yenikent-2020-survey,khan2005personality} and practical applications~\citep{bagby2016role} .

However, MBTI personality detection tends to be overly optimistic, as it currently does not align well with the natural distribution of personality traits within the population.
(1) Given the widespread use of MBTI, existing datasets are often derived from social media posts, with labels typically sourced from users' self-reported MBTI types~\citep{pandora}.
Assessments conducted by non-professional psychological institutions, along with inaccurate self-perception, can lead to \textbf{a mismatch between the self-reported personality traits and the actual linguistic patterns in the text}~\citep{McDonald2008MeasuringPC,Mller2019ControllingFR,Paulhus2007TheSM}.
For example, a user who self-identifies as an \textit{Extraversion} type might exhibit more \textit{Introversion} traits in their posts (Figure~\ref{fig:intro} (a)).
(2) From a psychological standpoint, \textbf{personality is not binary but rather complex, nuanced, and highly individualized}~\citep{Wu2022EstimatingTU}. 
Most people don't display extreme personality traits; instead, they tend to fall somewhere in the middle~\citep{Tzeng1989MeasurementAU}.
However, existing datasets use only binary MBTI labels, missing the full spectrum of personality traits in most people~\citep{Harvey1994ScoringTM,Wu2022EstimatingTU} (Figure~\ref{fig:intro} (b)).

\begin{wrapfigure}{r}{0.5\textwidth}
	% \vspace{-5mm}
    \centering
	\includegraphics[width=0.50\textwidth]{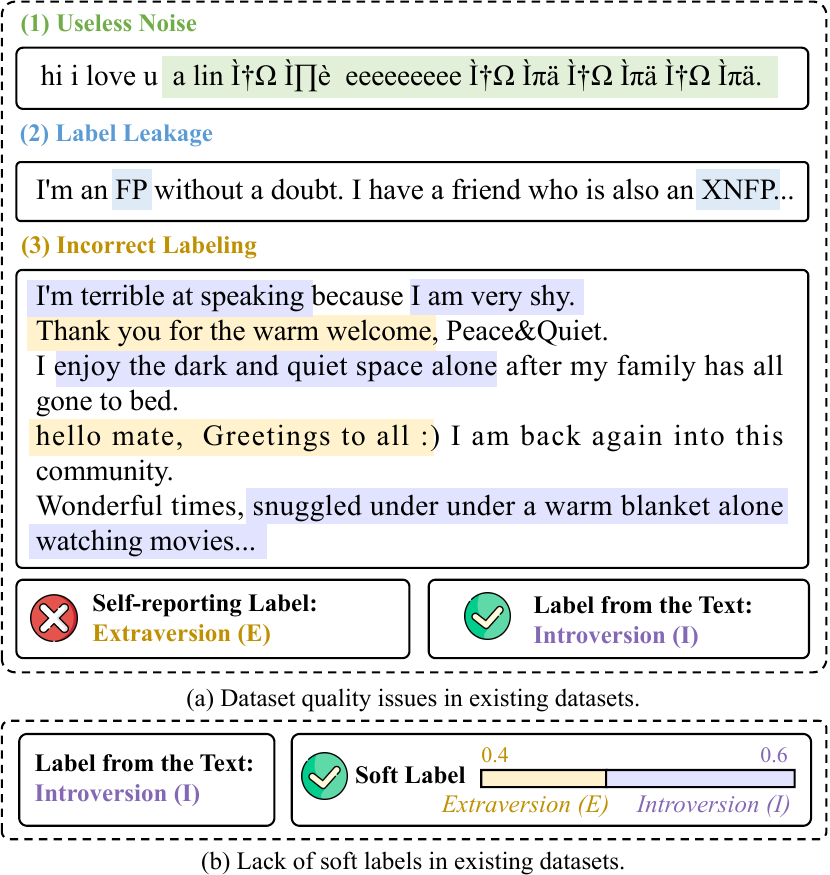}
    \caption{
    Our \ourdataset focuses on the above two limitations in existing MBTI personality detection datasets: data quality issues and the lack of soft labels.
    }
    % \vspace{-0.5cm}
\label{fig:intro}
\end{wrapfigure}

In this paper, we make a step toward solving the
challenge.
We optimize the task by constructing ~\ourdataset, the first MBTI personality detection dataset aligned with population traits.
(1) To solve the data quality issues related to self-reported labels, 
we propose the first data filtering guidelines for MBTI personality detection and apply them on existing datasets to ensoure data quality.
We manually re-annotate the cleaned datasets under the guidance of psychological experts, aligning each post with correct labels that best descrtibe the personality polarity.
(2) 
To capture the full range of population personality traits, we estimate soft labels for MBTI personality detection by deriving the polarity tendencies of samples.
The obtained soft labels confirm the above opinion that there are more people with non-extreme personality traits.

We analyze our \ourdataset in detail from multiple perspectives to explore the influence of soft labels compared to hard labels.
We further evaluate six large language models (LLMs) and four prompting methods on \ourdataset, and highlight the polarized predictions and biases as key directions.

Our contributions are as follows:
\begin{itemize}

 \item We are the first to challenge the over-optimism in MBTI personality detection from the nature of population personality traits.
  
  \item We optimize MBTI personality detection by creating \ourdataset, the first soft-labeled MBTI dataset, manually annotated under the guidance of psychologists.
 \item We highlight the polarized predictions and biases in LLMs as future directions. Our experimental results also confirm that soft labels can provide more benefits to other psychological tasks than hard labels.

\end{itemize}

\section{MBTI Personality Detection}
    
\begin{wraptable}{r}{0.4\textwidth}
\vspace{-2.5cm}
% \begin{adjustbox}{width=0.4\columnwidth,center}
\begin{adjustbox}{width=0.4\columnwidth}
        \begin{tabular}{l|cccc}
            \toprule
            Dataset & Types   & Train         & Validation   & \multicolumn{1}{c}{Test} \\ 
            \midrule
            \multirow{5}{*}{Kaggle}  & \textit{E/I}& $1194 / 4011 $& $409 / 1326$ & $396 / 1339$\\
            & \textit{S/N} & $727 / 4478$ & $222 / 1513$ & $248 / 1487$ \\
            & \textit{T/F} & $2410 / 2795$ & $791 / 944$ & $780 / 955$ \\
            & \textit{P/J} & $3096 / 2109$ & $1063 / 672$ & $1082 / 653$ \\
            \midrule
            \multirow{5}{*}{PANDORA} & \textit{E/I}& $1162 / 4278$& $386 / 1427$ & $377 / 1437$\\
            & \textit{S/N} & $610 / 4830$ & $208 / 1605$ & $210 / 1604$ \\
            & \textit{T/F} & $3549 / 1891$ & $1120 / 693$ & $1182 / 632$ \\
            & \textit{P/J} & $3211 / 2229$ & $1043 / 770$ & $1056 / 758$ \\
            \midrule
            \multirow{5}{*}{Twitter} & \textit{E/I}& $329 / 571$ & $99 / 201$ & $111 / 189$    \\
            & \textit{S/N} & $206 / 694$  & $66 / 234$ & $66 / 234$ \\
            & \textit{T/F} & $355 / 545$  & $139 / 161$ & $130 / 170$ \\
            & \textit{P/J} & $395 / 505$ & $112 / 188$ & $111 / 189$  \\ 
            \bottomrule
        \end{tabular}
        \end{adjustbox}
         \caption{Statistics of the Kaggle, PANDORA, and Twitter datasets.}
        \label{tab:dataset}
        \vspace{-1.5cm}
\end{wraptable}

\subsection{Existing Datasets Overview}

There are three widely used datasets in personality detection based on MBTI, including Twitter~\citep{Plank2015PersonalityTO}, Kaggle\footnote{\url{https://www.kaggle.com/datasnaek/mbti-type}}, PANDORA~\citep{pandora} (Table~\ref{tab:dataset}).
These three datasets are collected from social media posts. 
Each sample is a set of posts from the corresponding user, and the labels are obtained through users self-reporting their posts.
However, as shown in Figure~\ref{fig:intro}, there are issues of label leakage and irrelevant noise that significantly compromise the quality and factual accuracy of the data itself.
Moreover, users' lack of clarity in self-reporting often leads to mismatches between self-reported labels and their posts~\citep{McDonald2008MeasuringPC}.
To address the above issues, we clean and re-annotate existing datasets to construct our \ourdataset.

\begin{figure*}[!ht]
	\centering
    \vspace{0.4cm}
	\includegraphics[width=1.0\linewidth]{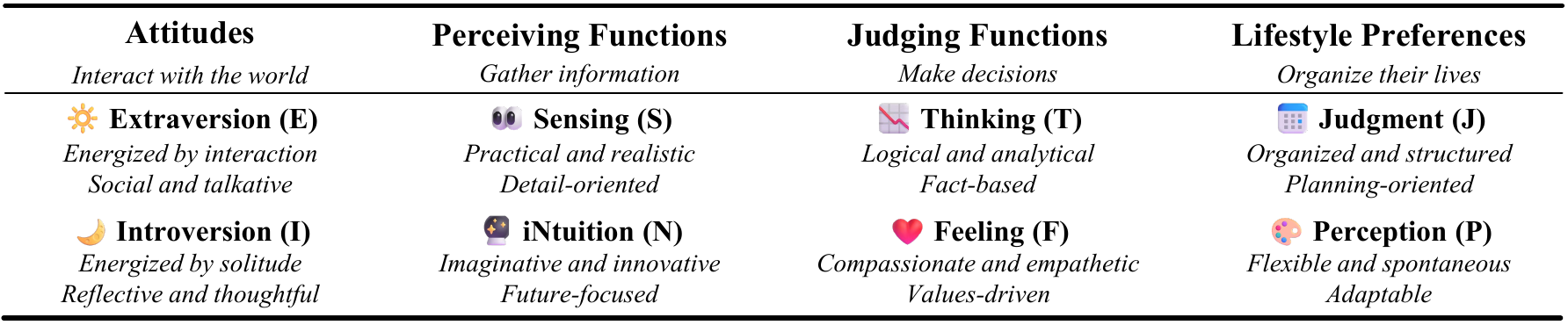}
    \caption{
Definitions of the four dimensions in MBTI personality theory.    }
    \vspace{-0.5cm}
\label{fig:mbti-def}
\end{figure*}

\subsection{Problem Overivew}
MBTI lays out a binary classification based on four distinct functions, and draws the typology of the person according to the combination of those four values (Figure~\ref{fig:mbti-def}):
\begin{itemize}
    \item Extraversion/Introversion (\textit{E/I}) - preference for how people direct and receive their energy, based on the outer or inner world.
    \item Sensing/INtuition (\textit{S/N}) - preference for how people take information in, by five senses or by interpretation and meanings.
    \item Thinking/Feeling (\textit{T/F}) - preference for how people make decisions, by relying on logic or emotions towards people and special circumstances.
    \item Judgment /Perception (\textit{J/P}) - how people deal with the world, by organizing it or staying open for new information.

\end{itemize}

The task of MBTI personality detection, which involves automatically inferring a person's MBTI type from their textual content, has attracted significant interest from researchers due to its broad range of potential applications~\citep{khan2005personality, bagby2016role}.
Given the growing interest in understanding personality through text, MBTI personality detection, that automatically inferring an individual's MBTI type from their written content, has become a prominent area of research, with a broad range of practical applications~\citep{khan2005personality, bagby2016role}.

As for hard labels, existing MBTI datasets lay out a binary classification based on four distinct dimensions independently (\textit{E/I}, \textit{S/N}, \textit{T/F}, \textit{J/P}), and draw the typology of the person according to the combination of those four values (e.g. \textit{ESTJ}).

In this paper, we construct a new dataset called \textsc{MbtiEval} with soft labels.
Soft labels are continual representations of polarity tendency mapped in $[0, 1]$, and we define the four dimensions of soft labels as the degree of \textit{E}, \textit{S}, \textit{T}, and \textit{J} polarity, respectively. 
For example, 40\% \textit{Extraverslon} simultaneously represents 60\% \textit{Introversion}.

\section{Dataset Construction}
    
In this paper, we re-annotate three existing datasets to solve the dataset quality issues. We introduce the construction details below.

\subsection{Design Principles}
\label{sec:design-priciples}

To ensure our annotation quality, we respectively establish data filtering guidelines for useless noise and label leakage issues, and data annotation guidelines for incorrect labeling issues.

\subsubsection{Data Filtering Guidelines}

We are the first to summarize data filtering guidelines for label leakage and useless noise issues in MBTI personality detection.

\begin{wrapfigure}{r}{0.5\textwidth}
	\vspace{-5mm}
    \centering
	\includegraphics[width=0.50\textwidth]{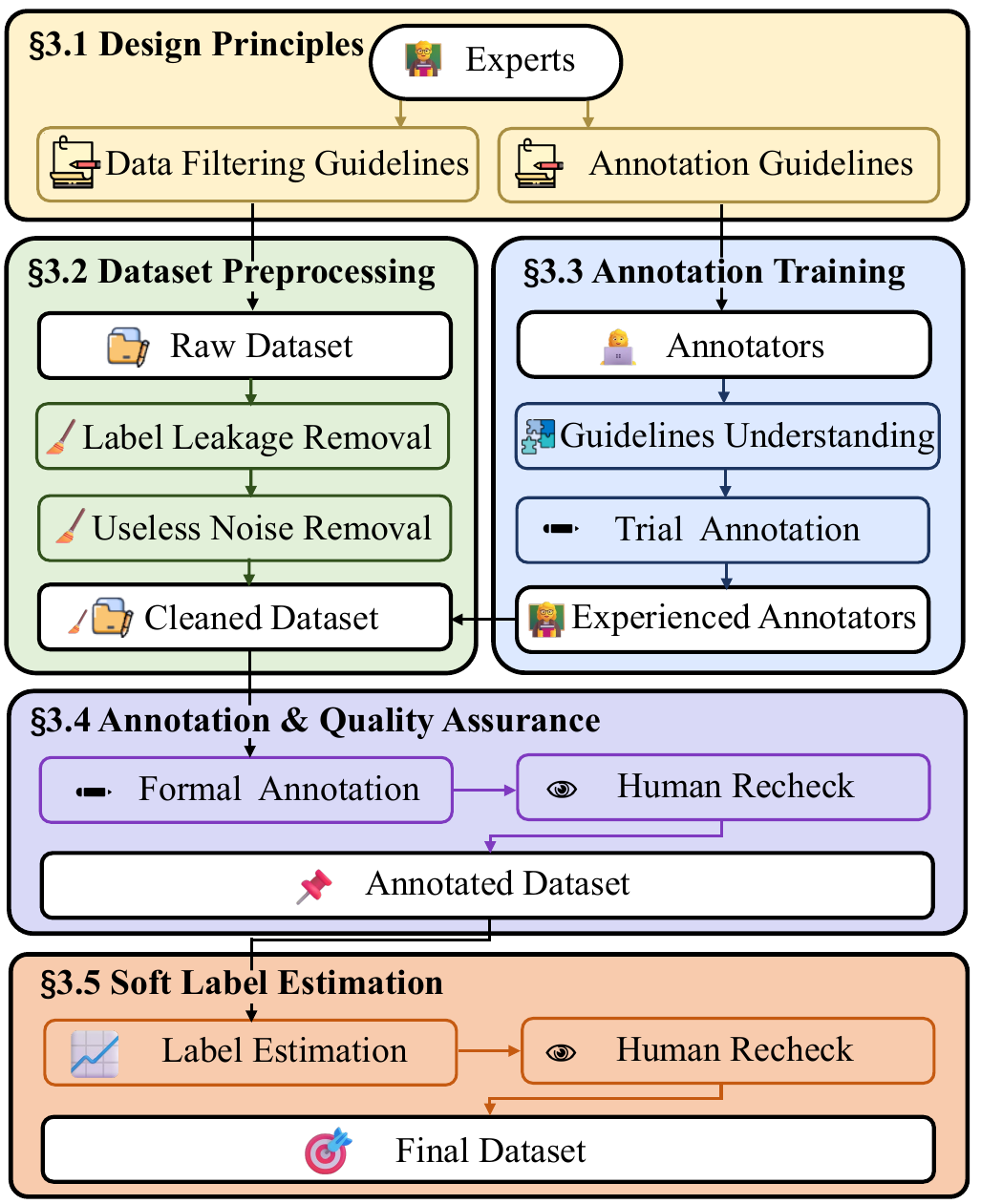}
    \caption{\ourdataset construction workflow.}
        % \vspace{-2cm}
	\label{fig:annotation}
\end{wrapfigure}

We divide the label leakage errors into three categories:  
(1) Direct Personality Leakage involves direct references to personality types through specific letters (``ENFP") or complete personality type words (``Introverted").
(2) Personality Trait Leakage involves specific MBTI trait descriptors providing enough context to infer certain personality types. For example, \textit{Te} indicates \textit{\underline{e}xtroverted \underline{T}hinking}. 
(3) Cross-Theory Trait Leakage involves traits from other personality theories like the Big Five~\citep{Furnham1996TheBF}, confirming MBTI personality characteristics.

We divide the useless noise issues into three categories:
(1) Information-insufficient Samples indicate samples that are too short (less than 100 words) to contain valuable information for inferring personality types. 
(2) Garbled Text indicates blocks of text that appears as random, unintelligible characters and symbols. 
(3) Link and Media References indicates hyperlinks or media file names, which are useless for personality detection.

\subsubsection{Data Annotation Guidelines}
\label{annoguide}
Personality refers to the combination of characteristics or qualities that form an individual's distinctive character~\citep{Mairesse2007UsingLC}.
It encompasses a wide range of traits, behaviors, thoughts, and emotional patterns that evolve from biological and environmental factors~\citep{Furnham1996TheBF,Celli2018IsBF,chung2011psychological,Line1948DescriptionAM}.
A particular personality can determine various outward observable properties or features, including consistent behavioral patterns, communication style, emotional expression and so on.

To address incorrect labeling issues and accurately label personality types,we refer to the methodology outlined in \citet{tajner2021WhyIM_EACL} to construct our annotation guidelines.
Psychology PhD students participate in the formulation of these guidelines. 
We discuss the personality traits for each dimension based on the dataset, analyzing and adjusting our guidelines through trial annotations.
Finally, we annotate the trial samples, which serve as expert guidelines.

\begin{figure*}[!ht]
	\centering
    \vspace{-0.2cm}
	\includegraphics[width=1.0\linewidth]{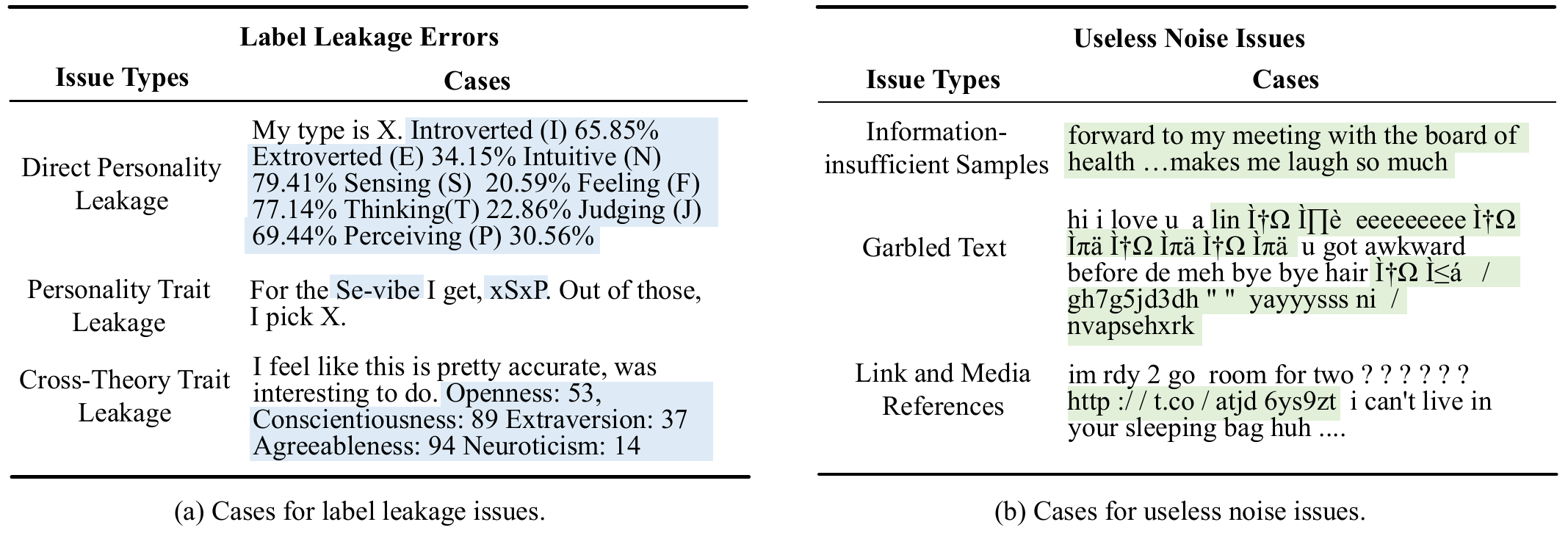}
    \caption{
We are the first to provide dataset quality issues annotation guidelines for MBTI personality detection.}
    \vspace{-0.5cm}
\label{fig:dataset-quality-guideline}
\end{figure*}

\subsection{Dataset Preprocessing}

We reconstruct the test sets of the three most commonly used personality detection datasets, including Twitter, Kaggle, and PANDORA. 
Following \citet{tajner2021WhyIM_EACL}, 
we select six samples for each type, totaling 286 samples across the three datasets.
\footnote{There are only four samples for the ESFJ type in the test set of PANDORA.}
To ensure that the samples provide useful data for personality detection, we filter the dataset. Specifically, we manually remove sentences involving label leakage and useless noise from the samples, using the annotation guidelines from Section~\ref{sec:design-priciples}, resulting in a cleaned dataset awaiting annotation.

\subsection{Annotation Training}
\label{sec:annotation-training}

\subsubsection{Annotation Guidelines Understanding}
We employ three experienced annotators who hold either a Bachelor's or Master's degree in English and can use English fluently with extensive annotation experience.
We conduct training sessions for them under the guidance of psychology experts, using expert-annotated examples to explain the definition and judgment signals for each MBTI dimension.
Annotators read through the entire instance and make independent judgments for each dimension.
We refer to the 4-point Likert scale ~\citep{Likert1932ATF} and ask the annotators to assign, for each instance and each MBTI dimension separately, two polar intensity labels.
For example, in the \textit{E/I} dimension, they could assign \textit{E-}, \textit{E+}, \textit{I-}, or \textit{I+} labels to reflect the degree of classification signals or their confidence level. 
In each pilot round, the dataset they annotate consists of 16 instances, which are not used in the final round to ensure the integrity of the final data.

\begin{wraptable}{r}{0.5\textwidth}
    \centering
    \setlength{\tabcolsep}{10pt}
    \begin{adjustbox}{width=0.50\textwidth,center}
    
    \begin{tabular}{ccllllll}
    \toprule
    % \multicolumn{2}{c}{} & \multicolumn{2}{c}{\textbf{Annotator 1}} & \multicolumn{2}{c}{\textbf{Annotator 2}} & \multicolumn{2}{c}{\textbf{Annotator 3}} \\
    \multicolumn{2}{c}{} & \multicolumn{2}{c}{\textbf{A1}} & \multicolumn{2}{c}{\textbf{A2}} & \multicolumn{2}{c}{\textbf{A3}} \\
    \midrule
    \multirow{5}[2]{*}{\textbf{E/I}} & OVERALL &       & 87.41\% &       & 87.76\% &       & 86.36\% \\
          & E+    &       & 92.00\% &       & 94.34\% &       & 100.00\% \\
          & E-    &       & 86.01\% \textcolor{arrowred}{↓} &       & 86.79\% &       & 86.90\% \\
          & I-    &       & 87.63\% &       & 84.62\% \textcolor{arrowred}{↓} &       & 83.47\% \textcolor{arrowred}{↓} \\
          & I+    &       & 90.48\% &       & 91.30\% &       & 100.00\% \\
    \midrule
    \multirow{5}[2]{*}{\textbf{S/N}} & OVERALL &       & 89.86\% &       & 85.31\% &       & 85.31\% \\
          & S+    &       & 87.93\% \textcolor{arrowred}{↓} &       & 86.57\% &       & 91.67\% \\
          & S-    &       & 88.12\% &       & 85.54\% &       & 81.45\% \textcolor{arrowred}{↓} \\
          & N-    &       & 90.48\% &       & 83.52\% \textcolor{arrowred}{↓} &       & 84.00\% \\
          & N+    &       & 100.00\% &       & 86.67\% &       & 100.00\% \\
    \midrule
    \multirow{5}[2]{*}{\textbf{T/F}} & OVERALL &       & 88.46\% &       & 89.86\% &       & 88.11\% \\
          & T+    &       & 93.33\% &       & 90.14\% &       & 94.29\% \\
          & T-    &       & 87.27\% &       & 88.73\% \textcolor{arrowred}{↓} &       & 90.43\% \\
          & F-    &       & 89.60\% &       & 90.24\% &       & 85.25\% \textcolor{arrowred}{↓} \\
          & F+    &       & 80.95\% \textcolor{arrowred}{↓} &       & 90.32\% &       & 85.71\% \\
    \midrule
    \multirow{5}[2]{*}{\textbf{J/P}} & OVERALL &       & 88.81\% &       & 87.41\% &       & 83.57\% \\
          & J+    &       & 89.66\% &       & 87.04\% &       & 95.83\% \\
          & J-    &       & 85.84\% \textcolor{arrowred}{↓} &       & 85.71\% \textcolor{arrowred}{↓} &       & 83.17\% \\
          & P-    &       & 91.20\% &       & 88.31\% &       & 83.33\% \\
          & P+    &       & 89.47\% &       & 88.73\% &       & 63.64\% \textcolor{arrowred}{↓} \\
    \bottomrule
    \end{tabular}%
    \end{adjustbox}
  \caption{The annotation accuracy of annotators across the four dimensions.}
  \label{tab:annotator-acc}%
    % \vspace{-0.4cm}
\end{wraptable}

\subsubsection{Trial Annotation}
We distribute trial annotation samples to the annotators, requiring them to annotate the same samples independently.
This is to assess their consistency and understanding of the guidelines, as well as the effectiveness of the guidelines themselves.
The trial annotation samples obtained during the pilot rounds show how the proposed guidelines are used in practice and highlight the most challenging aspects of the annotation process~\citep{shi-etal-2023-midmed,chen-etal-2023-automatic}. 

After the trial annotation, we determine the final result for each label by voting and measuring annotation quality using annotation accuracy and Fleiss' Kappa ~\citep{Fleiss1971MeasuringNS}. 
If the accuracy is below 0.8 or the Fleiss' Kappa does not exceed 0.45, we repeat the "Annotation Training" process, starting again from "Annotation Guidelines Understanding" to ensure the effectiveness of the trial annotation.\footnote{A Fleiss' Kappa value above 0.4 is satisfactory for subjective tasks~\citep{Jiang2019AutomaticTP}.}
We conduct three rounds of trial annotation, during which the accuracy and Kappa for the four dimensions steadily increase.

\subsection{Annotation and Quality Assurance}

\subsubsection{Formal Annotation}
Once the trial annotation is completed, the annotators become fully acquainted with the annotation guidelines and procedures.
The annotators are provided with the cleaned dataset and are expected to annotate independently. 
We pay them at a rate that is no less than the local average, and they are entitled to take adequate breaks to mitigate the effects of fatigue.
Annotators are provided with the cleaned dataset and annotate independently. They are compensated at a rate no less than the local average and are given adequate breaks to reduce fatigue.

\subsubsection{Human Recheck}

After the formal annotation, we conduct a human recheck of each annotated instance to ensure the annotation quality. 
If an instance is found to be unreasonably annotated, we summarize the issues and communicate them with the respective annotator.
These problematic samples are then mixed with other normal samples and re-annotated by the annotators to prevent direct suggestions or interference.
The re-annotated instances account for less than 15\% of the total.
The final Fleiss' Kappa scores\footnote{0.4779 for \textit{E/I}, 0.4686 for \textit{S/N}, 0.5517 for \textit{T/F}, 0.4622 for \textit{J/P}.} are not only satisfactory for personality detection~\citep{Jiang2019AutomaticTP} but also suitable for soft label estimation, which demonstrates our annotation quality.

\begin{algorithm}
\small
\caption{Soft Label Estimation for the \textit{E/I} Dimension}
\label{algorithm}
\begin{algorithmic}[1]
    \STATE \textbf{Input:} Annotated dataset with labels by three annotators. 
    
    \STATE \textbf{Initial True Label Calculation}
    \STATE \textbf{Co-occurrence Matrix Calculation}

    \STATE \textbf{Transition Matrix Formation}
    \STATE \textbf{Iterative Calculation}
    \REPEAT
        \STATE Compute the \textbf{posterior probability of E} based on the transition matrix \( \mathbf{T} \)
        \STATE Update the transition matrix \( \mathbf{T} \) using the posterior probability of E. 
    \UNTIL{change in posterior probability of E \(\leq\) tolerance. }

    \STATE \textbf{Tendency Calculation}
    
    \STATE \textbf{Normalization}
    
    \STATE \textbf{Output:} Soft Labels.
    
\end{algorithmic}
\end{algorithm}

\subsection{Soft Label Generation}

\label{sec:soft-label-generation}

\subsubsection{Label Estimation}
Inspired by the Dawid-Skene model~\citep{Dawid1979MaximumLE}, we adopt the EM algorithm~\citep{Dempster1977MaximumLF} to derive the polarity tendency of samples. The algorithm effectively fits annotators of varying quality and comprehensively produces realistic rankings for each sample ~\citep{Passonneau2013TheBO}.
We estimate a more objective soft label by processing the hard labels provided by the three annotators, thereby accurately reflecting the comprehensive inclination of the data across various dimensions.
In this paper, we use the \textit{E/I} dimension as an example to introduce the algorithm for estimating soft labels (Algorithm~\ref{algorithm}).

\tif{1.Initial True Label Calculation: }First, we take the annotated data as the initial input and calculate the median of the three annotators' labels as the initial true label.

\tif{2.Co-occurrence Matrix Calculation: }Next, we compare each annotator's label with the initial true label to obtain a matrix $\mathbf{M} \in \mathcal{R}^{3 \times 4 \times 4}$ where $m_{ijk}$ represents the joint frequency of category j and category k in i-th annotator labels.

\tif{3.Transition Matrix Formation: } Using Bayes' theorem to merge E+ and E- into E, and store the result in a new matrix T:

    \[ \small
    T_{i0k} = \frac{(N_{i0k} \times P_{E+}) + (N_{i1k} \times P_{E-})}{P_E}.
    \]
   $P_{E+}$, $P_{E-}$represent the proportions of labels classified as E+,E-,and $P_E$ is the sum of them.

\tif{4.Iterative Calculation: }Calculate the initial posterior probability of E from the transition matrix, then update the matrix and recompute until changes are within a predefined tolerance:
        \[ \small
        P(E \mid r_1, r_2, r_3) = \frac{\prod_{i=1}^{3} P_{r_i0} \times P_E}{\sum_{j=0}^{1}\prod_{i=1}^{3} P_{r_ij} \times P_E}, 
        \]
where $r_i$ represents the label labeled by the i-th annotator.

\tif{5.Tendency Calculation: }Since the midpoint where posterior probability is 0.5, we accumulate and average values on both sides.

\tif{6. Normalization: } Finally, we normalize the cumulative frequencies on both sides of the midpoint. Each combination is assigned a final ratio between 0 and 1, reflecting the label's tendency. We define this ratio as the "soft label" of the type \textit{Extraversion}.

\subsubsection{Human Recheck} 
Under the guidance of psychology experts, we review the estimated soft labels to ensure they align with the degree of personality polarity reflected in texts.
We find that the use of the EM algorithm in estimating soft labels effectively integrates the annotators' varying accuracy and labeling habits, producing accurate soft labels~\citep{raykar2010learning} (Table~\ref{tab:annotator-acc}).

\begin{figure*}[!ht]
	\centering
    \vspace{0.4cm}
	\includegraphics[width=1.0\linewidth]{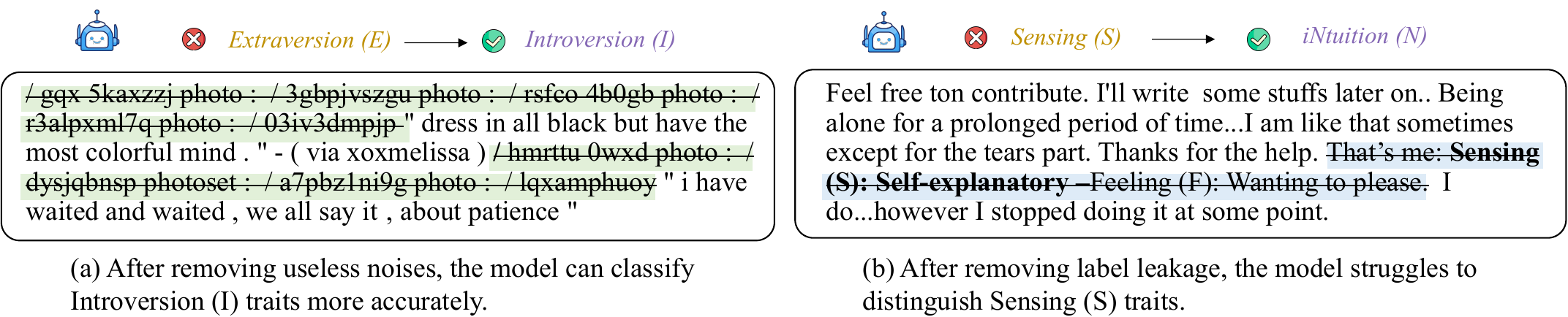}
    \caption{
After solving the data quality problem, the performance of the model is affected.    }
\label{fig:quality-case}
\end{figure*}

\section{Dataset Quality Analysis}
    We analyze the quality of our datasets from four perspectives to explore the influence of soft labels compared to hard labels~\citep{liu-etal-2020-towards-conversational}.

\begin{wraptable}{r}{0.5\textwidth}
    \centering
    \setlength{\tabcolsep}{10pt}
    \vspace{-0.5cm}
    \begin{adjustbox}{width=0.50\textwidth,center}
        \begin{tabular}{l|rrr}
            \toprule
            \textbf{Dimensions} & \textbf{Kaggle} & \textbf{Twitter} & \textbf{PANDORA} \\ 
            \midrule
            Useless Noise & 15.92\% & 43.13\% & 2.65\% \\
            Label Leakage & 31.21\% & 0.62\% & 16.56\% \\
            Incorrect Labeling & 29.17\% & 29.17\% & 29.79\% \\
            \bottomrule
        \end{tabular}
    \end{adjustbox}
    \caption{Issues solved in three datasets.}
    \vspace{-1cm}
    \label{tab:issues-in-dataset}
\end{wraptable}

\subsection{Self-reported Labels Influence Model Performance}

We analyze the useless noise issues, self-reported label leakage, and incorrect self-reported labels in Table~\ref{tab:issues-in-dataset} and Figure~\ref{fig:quality-case}.
The datasets exhibit varying degrees of quality issues before refinement that impact model performance.
Therefore, filtering the input data to avoid such influence is crucial and we are the first to establish guidelines for manual filtering (Section~\ref{sec:design-priciples}).

\begin{figure*}[!ht]
	\centering
    \vspace{0.4cm}
	\includegraphics[width=1.0\linewidth]{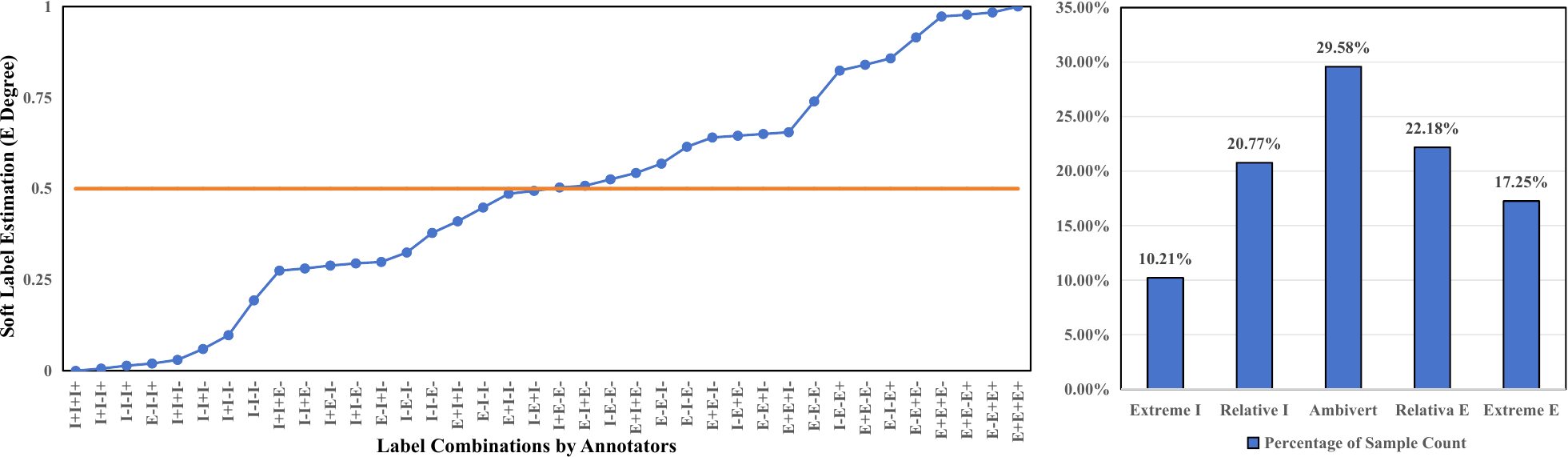}
    \caption{
    The mapping from hard labels to soft labels of the \textit{E/I} dimension in
\ourdataset. In (a), we use an EM algorithm to convert annotator label combinations (x-axis) into soft labels ranging from 0 to 1 (y-axis). We then define Extreme I, Relative I, Ambivert, Relative E, and Extreme E based on these soft label values, as shown on the x-axis of (b). The y-axis in (b) shows the sample count for each category.
    }
\label{fig:IE}
\end{figure*}

\subsection{Soft Labels Align with Population Traits}

We illustrate the estimated soft labels and the corresponding sample count distribution of \textit{E/I}, \textit{S/N}, \textit{T/F}, and \textit{J/P} in Figure~\ref{fig:IE} and Appendix Figure~\ref{fig:SN}-\ref{fig:JP}.
Soft labels exhibit a smooth distribution from 0 to 1, and there are more non-extreme samples than extreme ones.
This is consistent with the population distribution and demonstrates that our soft labels align with population traits better compared to hard labels~\citep{Harvey1994ScoringTM}.

\subsection{Label Shifts Across Personality Dimensions}
\label{app-sec:Ensuring_Text-Label_Consistency}

\begin{wraptable}{r}{0.5\textwidth}
    \centering
    \setlength{\tabcolsep}{10pt}
    \begin{adjustbox}{width=0.50\textwidth,center}
            \begin{tabular}{c|ccc}
            \toprule
            Dimensions  & Twitter & Kaggle & PANDORA \\ 
            \midrule
            E / I & $55~\textcolor{arrowgreen}{\uparrow}~ / ~41~\textcolor{arrowred}{\downarrow}~ / ~\Delta=7$& $56~\textcolor{arrowgreen}{\uparrow}~ / ~40~\textcolor{arrowred}{\downarrow}~ / ~\Delta=8$  & $49~\textcolor{arrowgreen}{\uparrow}~ / ~45~\textcolor{arrowred}{\downarrow}~ / ~\Delta=3$ \\
            S / N  & $53~\textcolor{arrowgreen}{\uparrow}~ / ~43~\textcolor{arrowred}{\downarrow}~ / ~\Delta=5$ & $51~\textcolor{arrowgreen}{\uparrow}~ / ~45~\textcolor{arrowred}{\downarrow}~ / ~\Delta=3$& $46~\textcolor{black}{\textbf{--}}~ / ~48~\textcolor{black}{\textbf{--}}~ / ~\Delta=0$ \\
            T / F & $38~\textcolor{arrowred}{\downarrow}~ / ~58~\textcolor{arrowgreen}{\uparrow}~ / ~\Delta=10$ & $50~\textcolor{arrowgreen}{\uparrow}~ / ~46~\textcolor{arrowred}{\downarrow}~ / ~\Delta=2$ & $53~\textcolor{arrowgreen}{\uparrow}~ / ~41~\textcolor{arrowred}{\downarrow}~ / ~\Delta=5$ \\
            P / J & $53~\textcolor{arrowgreen}{\uparrow}~ / ~43~\textcolor{arrowred}{\downarrow}~ / ~\Delta=5$& $51~\textcolor{arrowgreen}{\uparrow}~ / ~45~\textcolor{arrowred}{\downarrow}~ / ~\Delta=3$  & $46~\textcolor{black}{\textbf{--}}~ / ~48~\textcolor{black}{\textbf{--}}~ / ~\Delta=0$ \\
            \bottomrule
        \end{tabular}
    \end{adjustbox}
    \caption{Statistics of three datasets, showing changes in comparison with baseline.}
    \label{tab:dataset-dimension-num}
    % \vspace{-1cm}
\end{wraptable}

We quantify label changes across four dimensions before and after refinement in the three datasets (Table~\ref{tab:dataset-dimension-num}).
Across all datasets, a notable shift from \textit{I} to \textit{E} is observed, likely reflecting social media's interactive nature that promotes extroverted (\textit{E}) behaviors.
The \textit{S/J} to \textit{N/P} shift aligns with the fragmented and immediate communication style typical of these platforms.
Regarding the \textit{T/F} dimension, Kaggle and PANDORA show more \textit{F} labels reclassified as \textit{T}, possibly due to the rational and in-depth exchanges prevalent on these platforms. 
In contrast, Twitter shows a greater shift from \textit{T} to \textit{F}, reflecting its focus on emotion-driven content.

\begin{figure*}[!ht]
	\centering
    \vspace{0.4cm}
	\includegraphics[width=1.0\linewidth]{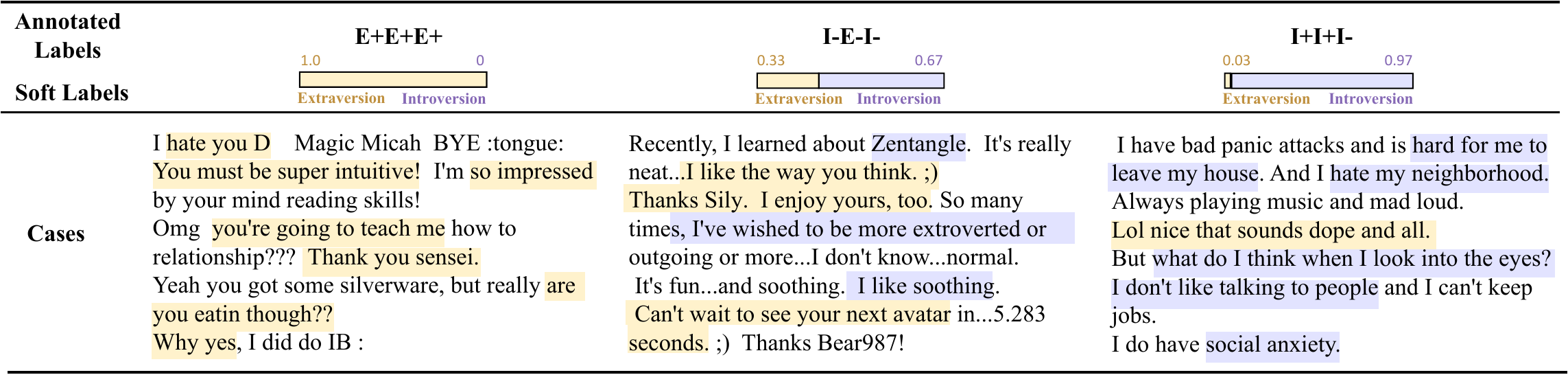}
    \caption{
    Three cases and their corresponding annotated labels and soft labels.
    }
\label{fig:soft-label-sample-case}
\end{figure*}

\subsection{Soft Labels Capture Personality Polarity Tendencies}
% \subsection{Validating Personality Polarity with Soft Labels}
\label{app-sec:Validating_Personality_Polarity}

We provide three case as well as their corresponding annotated labels and soft labels in Figure~\ref{fig:soft-label-sample-case}.
The personality tendencies displayed in user posts are well captured by the annotators and reflected in the annotated labels, while the estimated soft labels also demonstrate a tendency change from extraversion to introversion.

\section{Experiment Setup}
    To evaluate model performance on \ourdataset and explore future directions, we conduct experiments across six backbone models and four prompting methods on \ourdataset~\citep{liu-etal-2022-go}. In this section, we provide an overview of the experiments.

\subsection{Experimental Details}
\label{sec:bakcbone-prompting}

\subsubsection{Backbones}
\label{sec-app:Backbones}

We employ several widely used and powerful LLMs as our backbones, including closed-source LLMs: GPT-4~\citep{achiam2023gpt} (\texttt{gpt-4o-mini} and \texttt{gpt-4o}), and open-source LLMs: Qwen2 series~\citep{yang2024qwen2} (\texttt{Qwen2-7B-Instruct} and \texttt{Qwen2-72B-Instruct}), Llama3.1 series~\citep{dubey2024llama} (\texttt{Meta-Llama-3.1-8B-Instruct} and \texttt{Meta-Llama-3.1-70B-Instruct}).
Qwen2 series have a context window of 32K tokens, while Llama 3.1 series and GPT-4o series both have context windows of 128K tokens, which are sufficient to process a large volume of user posts and perform Chain-of-Thought (CoT) \citep{wei2022chain} reasoning.
Moreover, we employ the average soft labels of each dimension as the \textit{baseline}, to evaluate model performance.

\subsubsection{Prompting Methods}
\label{sec:prompting-approaches}
Following the methodologies outlined in \citet{Yang2023PsyCoTPQ}, based on each user's posts $P$, we apply them to an inference prompt template $t(\cdot)$, where $t(P)$ serves as the actual input to LLMs.
We adopt four prompting approaches:
(1) Zero-shot involves directly presenting the task description and requiring the model to directly complete the task.\footnote{\textit{Standard} in \citet{Yang2023PsyCoTPQ}.}
(2) Step-by-step~\citep{kojima2022large}
employs the additional phrase \textit{Let's think step by step} based on zero-shot in the inference prompt.\footnote{\textit{Zero-shot} in \citet{Yang2023PsyCoTPQ}.}
(3) Few-shot
employs two additional examples in the inference prompt based on zero-shot, with each example containing a post and the corresponding complete personality label.\footnote{We manually choose the two examples.}
(4) PsyCoT~\citep{Yang2023PsyCoTPQ}
requires the model to answer MBTI scale questions and refer to the answers for final personality judgments.
We present the detailed content of the prompting methods in Appendix Table~\ref{table:inference_template_softlabel}.
We set $temperature = 0$ to eliminate randomness. Each post is truncated to no more than 80 tokens.

\subsection{Evaluation}
Following \citet{Raykar2010LearningFC} and \citet{pandora}, we directly predict soft labels to evaluate the model's ability to make more detailed estimations of personality traits.
Specifically, $t(\cdot)$ instructs the model to assess the degree of each personality trait and generate a score within the range of 1 to 9, for example, \textit{[[7.25]]}.

We employ the Root Mean Square Error (RMSE) and Mean Absolute Error (MAE) to measure the differences between the predicted soft labels by LLMs and the estimated golden ones.
Considering the limited precision of our estimated soft labels, we introduce Segmented MAE (S-MAE) and Segmented RMSE (S-RMSE).
Instead of computing the error based solely on the raw predictions, we discretize the continuous 0-1 distribution into ten intervals and evaluate the corresponding RMSE and MAE within these segments.
By framing the problem in this way, the error becomes less sensitive to specific small deviations in continuous values and more focused on whether the prediction falls within an appropriate range.
Further details are introduced in Appendix~\ref{app-sec:metrics}.

\setlength{\tabcolsep}{1pt} % Adjust this value to reduce column spacing
\begin{table*}[!ht]
  \centering
\begin{adjustbox}{width=1.0\textwidth}
    % \small
    \footnotesize

\begin{tabular}{c@{\hspace{5pt}}c@{\hspace{8pt}}c@{\hspace{5pt}}c@{\hspace{8pt}}c@{\hspace{5pt}}c@{\hspace{8pt}}c@{\hspace{5pt}}c@{\hspace{8pt}}c@{\hspace{5pt}}c@{\hspace{8pt}}c}
    \toprule
    \multirow{2}[2]{*}{\textbf{Backbones}} & \multirow{2}[2]{*}{\textbf{Methods}} & \multicolumn{2}{c}{\textbf{E/I}} & \multicolumn{2}{c}{\textbf{S/N}} & \multicolumn{2}{c}{\textbf{T/F}} & \multicolumn{2}{c}{\textbf{J/P}} & \multirow{2}[2]{*}{\textbf{Rank}}                                                                 \\
                                           &                                      & \textbf{S-RMSE}                  & \textbf{S-MAE}                   & \textbf{S-RMSE}                  & \textbf{S-MAE}                   & \textbf{S-RMSE}                   & \textbf{S-MAE}   & \textbf{S-RMSE}  & \textbf{S-MAE}   &      \\
    \midrule
    Baseline                               & -                                    & 2.66                             & 2.29                             & 2.70                             & 2.31                             & 2.82                              & 2.43             & 3.04             & 2.59             & -    \\
    \midrule
    \multirow{4}[2]{*}{gpt-4o-mini}        & Zero-shot                            & \textbf{2.78}                    & \textbf{2.17}                    & \textbf{2.77}                    & \textbf{2.24}                    & \textbf{2.37}                     & \textbf{1.99}    & \textbf{3.11}             & \textbf{2.60}             & \textbf{1.00} \\
                                           & Step-by-step                         & \underline{2.82}                 & \underline{2.23}                 & 3.02                             & 2.42                             & 2.67                              & 2.13             & 3.52             & 2.89             & 2.63 \\
                                           & Few-shot                             & 2.96                             & 2.43                             & \underline{2.84}                 & \underline{2.41}                 & \underline{2.65}                  & \underline{2.15} & \underline{3.15}             & \textbf{2.60}             & \underline{2.38} \\
                                           & PsyCoT                               & 3.02                             & 2.34                             & 4.65                             & 3.90                             & 3.61                              & 2.87             & 3.96             & 3.12             & 3.88 \\
    \midrule
    \multirow{4}[2]{*}{gpt-4o}             & Zero-shot                            & \underline{2.91}                 & \underline{2.28}                 & \textbf{2.90}                    & \textbf{2.23}                    & \textbf{2.62}                     & \textbf{2.06}    & 3.80             & 3.15             & \textbf{1.75} \\
                                           & Step-by-step                         & 3.07                             & 2.38                             & \underline{2.96}                 & \underline{2.33}                 & \underline{2.73}                  & 2.19             & \underline{3.51}             & \underline{2.88}             & 2.38 \\
                                           & Few-shot                             & \textbf{2.82}                    & \textbf{2.13}                    & 3.34                             & 2.74                             & 2.76                              & \underline{2.16} & \textbf{3.44}             & \textbf{2.81}             & \underline{1.88} \\
                                           & PsyCoT                               & 3.36                             & 2.67                             & 5.02                             & 4.25                             & 3.23                              & 2.49             & 5.20             & 4.27             & 4.00 \\
    \midrule
    \multirow{4}[2]{*}{Qwen2-7B}           & Zero-shot                            & \underline{3.11}±\tiny{0.00}     & \underline{2.59}±\tiny{0.00}     & \textbf{2.68}±\tiny{0.01}        & \textbf{2.29}±\tiny{0.01}        & \underline{2.90}±\tiny{0.00}      & \underline{2.48}±\tiny{0.00} & \underline{3.16}±\tiny{0.02} & \underline{2.65}±\tiny{0.01} & \underline{1.75} \\
                                           & Step-by-step                         & 3.26±\tiny{0.08}                 & 2.70±\tiny{0.06}                 & \underline{2.69}±\tiny{0.00}     & \underline{2.30}±\tiny{0.00}     & 2.92±\tiny{0.00}                  & 2.50±\tiny{0.00} & \underline{3.16}±\tiny{0.00} & 2.66±\tiny{0.00} & 2.88 \\
                                           & Few-shot                             & 3.51±\tiny{0.01}                 & 2.80±\tiny{0.01}                 & 3.29±\tiny{0.01}                 & 2.71±\tiny{0.01}                 & 4.03±\tiny{0.04}                  & 3.23±\tiny{0.04} & 3.20±\tiny{0.02} & 2.68±\tiny{0.02} & 4.00 \\
                                           & PsyCoT                               & \textbf{2.68}±\tiny{0.00}        & \textbf{2.30}±\tiny{0.00}        & 2.70±\tiny{0.00}                 & \underline{2.30}±\tiny{0.01}     & \textbf{2.82}±\tiny{0.00}         & \textbf{2.43}±\tiny{0.00} & \textbf{3.10}±\tiny{0.01} & \textbf{2.62}±\tiny{0.01} & \textbf{1.38} \\
    \midrule
    \multirow{4}[2]{*}{Qwen2-72B}          & Zero-shot                            & \underline{2.58}±\tiny{0.00}     & \underline{2.11}±\tiny{0.00}     & \textbf{2.70}±\tiny{0.01}        & \textbf{2.28}±\tiny{0.01}        & \underline{3.00}±\tiny{0.02}      & \underline{2.46}±\tiny{0.02} & \underline{3.10}±\tiny{0.01} & \underline{2.61}±\tiny{0.01} & \textbf{1.75} \\
                                           & Step-by-step                         & \textbf{2.58}±\tiny{0.01}        & \textbf{2.10}±\tiny{0.01}        & \textbf{2.70}±\tiny{0.01}        & \underline{2.30}±\tiny{0.01}     & \textbf{2.92}±\tiny{0.01}         & \textbf{2.41}±\tiny{0.01} & 3.12±\tiny{0.02} & 2.62±\tiny{0.01} & \textbf{1.75} \\
                                           & Few-shot                             & 2.76±\tiny{0.02}                 & 2.26±\tiny{0.01}                 & 3.09±\tiny{0.02}                 & 2.56±\tiny{0.02}                 & 3.29±\tiny{0.03}                  & 2.70±\tiny{0.02} & \textbf{3.09}±\tiny{0.01} & \textbf{2.60}±\tiny{0.01} & 2.75 \\
                                           & PsyCoT                               & 2.64±\tiny{0.01}                 & 2.15±\tiny{0.01}                 & 3.99±\tiny{0.03}                 & 3.23±\tiny{0.03}                 & 4.59±\tiny{0.03}                  & 3.75±\tiny{0.03} & 4.66±\tiny{0.04} & 3.75±\tiny{0.04} & 3.75 \\
    \midrule
    \multirow{4}[2]{*}{Llama3.1-8B}        & Zero-shot                            & \textbf{2.77}±\tiny{0.02}        & \textbf{2.39}±\tiny{0.01}        & \textbf{2.83}±\tiny{0.01}        & \textbf{2.37}±\tiny{0.00}        & \textbf{2.86}±\tiny{0.02}         & \textbf{2.44}±\tiny{0.01} & \textbf{3.10}±\tiny{0.01} & \textbf{2.64}±\tiny{0.01} & \textbf{1.00} \\
                                           & Step-by-step                         & 3.76±\tiny{0.05}                 & 3.05±\tiny{0.03}                 & 3.82±\tiny{0.07}                 & 3.07±\tiny{0.08}                 & 3.93±\tiny{0.12}                  & 3.21±\tiny{0.13} & 3.73±\tiny{0.02} & 2.99±\tiny{0.03} & 3.88 \\
                                           & Few-shot                             & 3.67±\tiny{0.00}                 & 3.00±\tiny{0.01}                 & \underline{3.42}±\tiny{0.00}     & \underline{2.77}±\tiny{0.00}     & 3.66±\tiny{0.00}                  & 3.01±\tiny{0.00} & \underline{3.44}±\tiny{0.00} & \underline{2.87}±\tiny{0.00} & \underline{2.50} \\
                                           & PsyCoT                               & \underline{3.31}±\tiny{0.02}     & \underline{2.80}±\tiny{0.02}     & 3.55±\tiny{0.00}                 & 2.93±\tiny{0.01}                 & \underline{3.54}±\tiny{0.00}      & \underline{2.96}±\tiny{0.00} & 3.70±\tiny{0.01} & 3.12±\tiny{0.00} & 2.63 \\
    \midrule
    \multirow{4}[2]{*}{Llama3.1-70B}       & Zero-shot                            & \textbf{2.68}±\tiny{0.01}        & \textbf{2.22}±\tiny{0.00}        & \textbf{3.15}±\tiny{0.01}        & \underline{2.58}±\tiny{0.01}     & \underline{2.89}±\tiny{0.01}      & 2.41±\tiny{0.01} & \textbf{3.24}±\tiny{0.02} & \textbf{2.72}±\tiny{0.01} & \textbf{1.50} \\
                                           & Step-by-step                         & \underline{2.88}±\tiny{0.01}     & 2.39±\tiny{0.01}                 & 3.20±\tiny{0.01}                 & 2.63±\tiny{0.01}                 & 2.92±\tiny{0.02}                  & \underline{2.38}±\tiny{0.02} & 3.37±\tiny{0.02} & \underline{2.84}±\tiny{0.01} & 2.75 \\
                                           & Few-shot                             & 2.93±\tiny{0.02}                 & 2.47±\tiny{0.02}                 & \underline{3.17}±\tiny{0.02}     & \textbf{2.55}±\tiny{0.02}        & \textbf{2.75}±\tiny{0.01}         & \textbf{2.24}±\tiny{0.02} & \underline{3.25}±\tiny{0.02} & 2.78±\tiny{0.01} & \underline{2.00} \\
                                           & PsyCoT                               & 2.94±\tiny{0.02}                 & \underline{2.34}±\tiny{0.01}     & 3.96±\tiny{0.01}                 & 3.17±\tiny{0.01}                 & 3.65±\tiny{0.03}                  & 2.93±\tiny{0.03} & 4.00±\tiny{0.02} & 3.19±\tiny{0.03} & 3.75 \\
    \bottomrule
\end{tabular}%

  \end{adjustbox}
  \caption{We conduct experiments using soft labels and use S-RMSE and S-MAE as evaluation metrics, where lower values indicate better performance. The best and second-best results across the four methods for each backbone are \textbf{bolded} and \underline{underlined}.}
  % \vspace{-0.4cm}
  \label{tab:eval_softlabel_pcc_ce}%
\end{table*}%

\section{Results and Analysis}
    
\subsection{Model Evaluation}
\label{sec:model-evaluation}
\begin{figure*}[!ht]
	\centering
    \vspace{0.4cm}
	\includegraphics[width=1.0\linewidth]{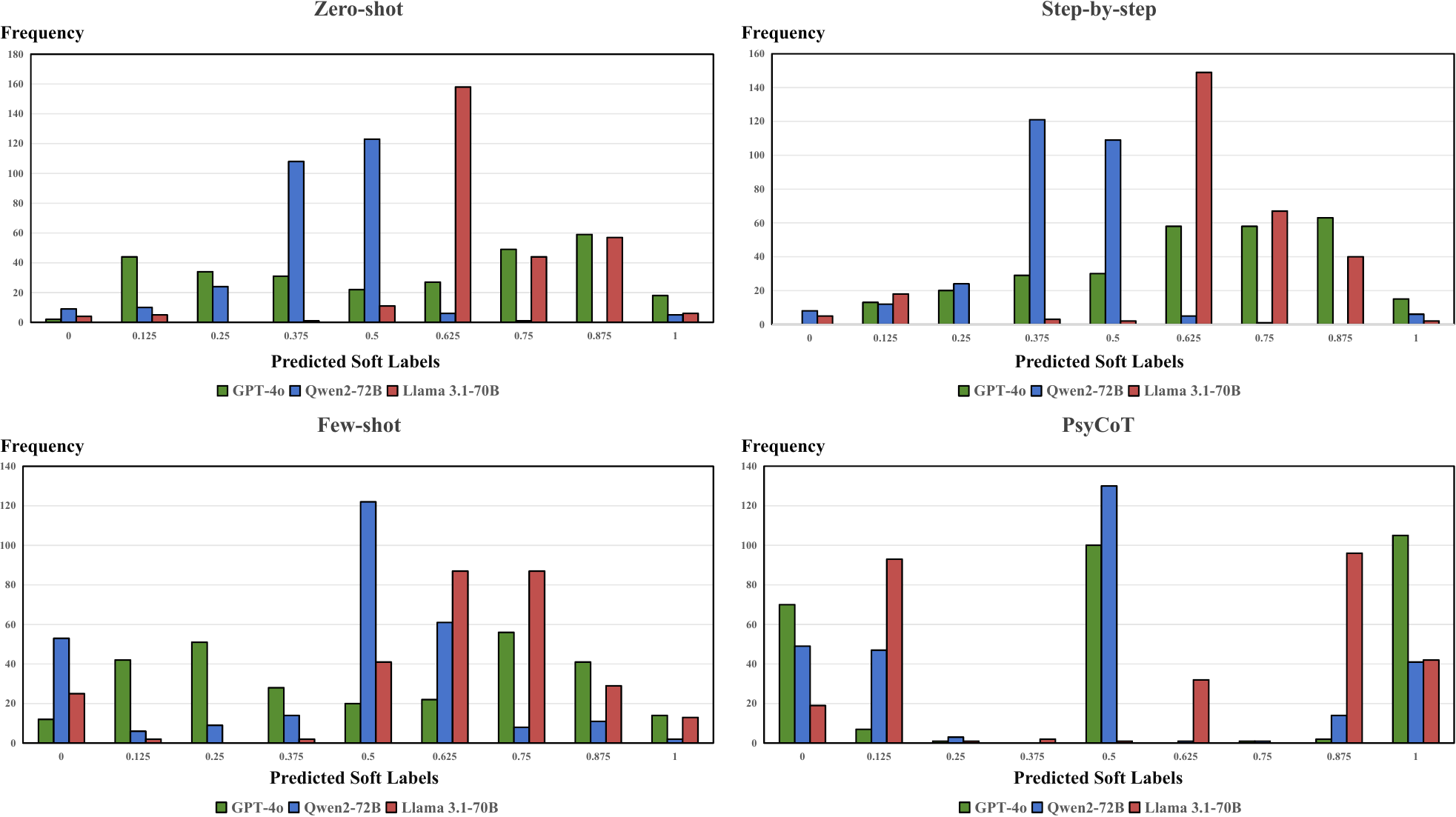}
    \caption{The score distribution of LLMs on \ourdataset for the \textit{T/F} dimension.}
    \vspace{-0.4cm}
\label{fig:finalscore}
\end{figure*}

In this paper, we conduct experiments on \ourdataset across four inference prompt methods and six backbones introduced in Section~\ref{sec:bakcbone-prompting}. The main results of our experiments are shown in Table~\ref{tab:eval_softlabel_pcc_ce} and Figure~\ref{fig:finalscore}. We address the following research questions (RQs):

\paragraph{RQ 1: Can LLMs consistently outperform simple baselines in soft label prediction?}
We provide a baseline model introduced in Sec.~\ref{sec-app:Backbones}, which computes the mean soft label for each MBTI dimension across the entire dataset, and evaluate several LLMs using different prompting methods. Surprisingly, in certain cases, even the best prompting methods for these LLMs fail to outperform the baseline model.
For instance, the baseline model achieves an S-RMSE of 3.722 and an S-MAE of 3.017 for the \textit{S/N} dimension.
However, the few-shot prompting method using the Llama3.1-70B backbone only achieved an S-RMSE of 3.754 and an S-MAE of 3.073, which is worse than the baseline. 
These results suggest that for certain models, 
the model may struggle to capture the underlying personality distribution as effectively as a simple baseline that leverages the dataset's overall distribution.

\paragraph{RQ 2: How do different methods perform on \ourdataset?} The Zero-shot method demonstrates superior overall performance across all six backbones.
As shown in Table~\ref{tab:eval_softlabel_pcc_ce}, it achieves the lowest average S-MAE. % (Rank 1).
This means that the Zero-shot method performs well on the majority of samples, with relatively small errors and the lowest average error.
Zero-shot achieves the lowest average RMSE on most backbones, and achieves comparable RMSE with the best method on gpt-4o and Llama3.1-8B.
Considering that S-RMSE is more sensitive to larger errors (i.e., cases with extreme deviations in predictions), this indicates that the zero-shot model performs well overall, but may have significant deviations in the extreme values of certain backbones.
In contrast, other methods may amplify the inherent biases of the models. For example, the PsyCoT method tends to predict 0 in certain dimensions, while few-shot methods are more influenced by the provided examples.

\paragraph{RQ 3: How do different backbone models exhibit biases in soft label prediction?} 
Different backbone models exhibit varying biases in their soft label predictions (Figure~\ref{fig:finalscore} and Appendix Figure~\ref{fig:finalscoreei}-~\ref{fig:finalscorejp}).\footnote{We show the results of each first round on three backbones.}
GPT-4o shows less bias compared to Llama 3.1, while Qwen2 more frequently assigns a score of 5. We hypothesize that these discrepancies stem from differences in their training corpora, leading to varied score distributions.

\subsection{Discussion}
\subsubsection{Can Soft Labels Better Align with Population Traits?}

\begin{wrapfigure}{r}{0.5\textwidth}
	\vspace{-5mm}
    \centering
	\includegraphics[width=0.45\textwidth]{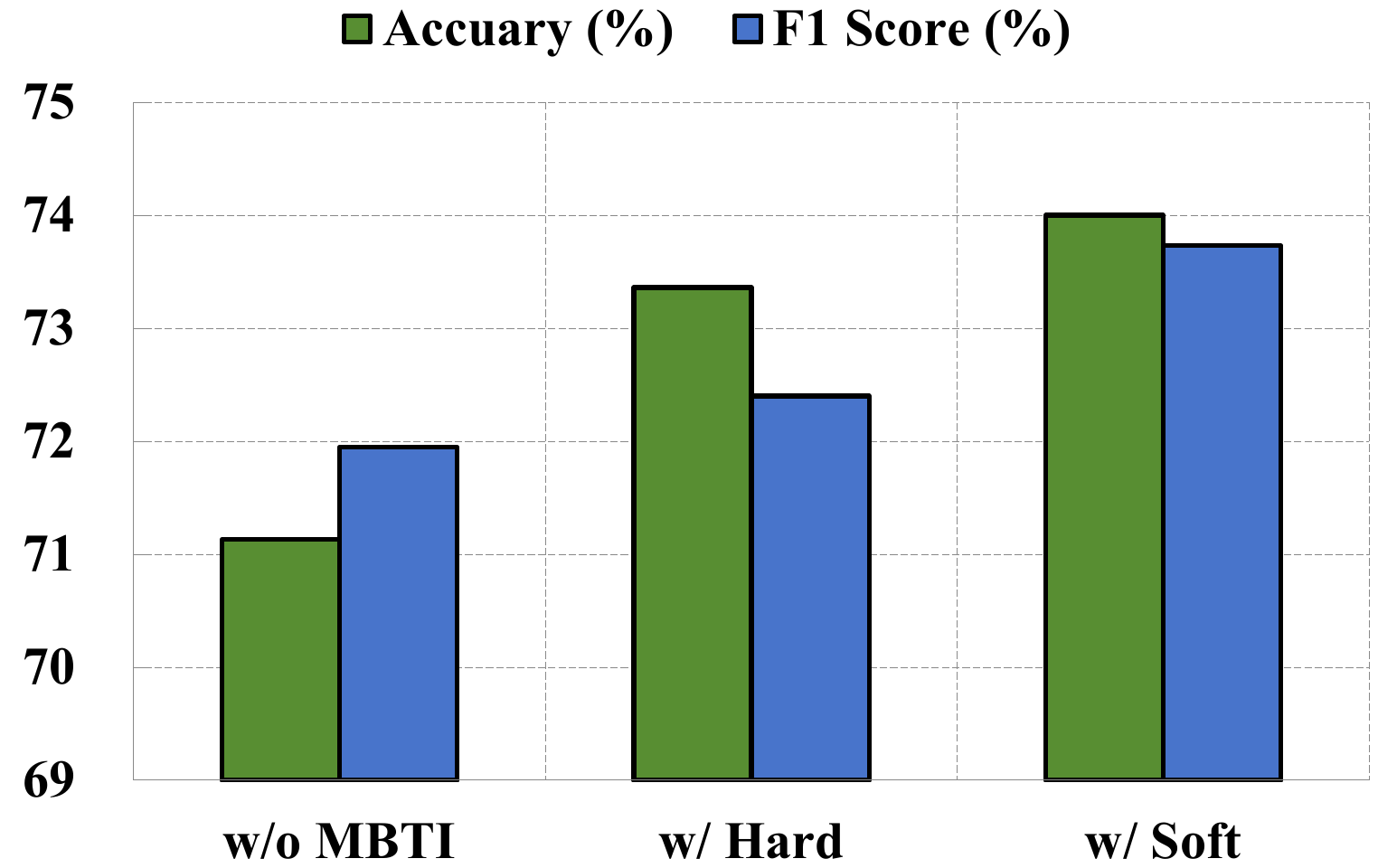}
    \caption{We integrate personality traits into stress detection, demonstrating that the use of soft labels (\textit{w/ Soft}) enhances performance on personality-related tasks~.}
    \label{fig:procon_result}
	\vspace{-0.2cm}
\label{fig:intro}
\end{wrapfigure}

To validate the effectiveness of soft labels, we 
follow \citet{Ji2023IsCA} and 
incorporat personality traits into the stress identification task.
This task is both challenging and representative, with substantial real-world significance.
We use the Dreaddit stress detection dataset~\citep{turcan2019dreaddit}, where the LLM is provided with a \textit{post} from one poster and tasked with determining whether the poster is likely to suffer from very severe stress or not (\textit{w/o MBTI}).
We predict the user's MBTI personality types using both hard-label (\textit{w/ Hard}) and soft-label (\textit{w/ Soft}) approaches with LLM, based on the \textit{post}. These personality types are then used as auxiliary inputs for stress identification. To evaluate the robustness of our results, we repeat the stress identification experiments 10 times and perform a t-test to assess the statistical significance of the outcomes.
The results are shown in Figure~\ref{fig:procon_result}.

We analyze the experimental results conclude that: (1) The inferior performance of \textit{w/o MBTI} (Acc: 72.13, F1: 71.95) indicates that MBTI personality traits contribute to stress identification. 
(2) \textit{w/ Soft} (Acc: 74.00, F1: 73.73) outperforms \textit{w/ Hard} (Acc: 73.36, F1: 72.40), indicating that soft labels more accurately align with population personality traits than hard labels.

\subsubsection{What Is the Difference in Behavior Between Hard Labels and Soft Labels?}

We evaluate the performance on \ourdataset using the mapped hard labels and analyze the distribution of both hard and soft labels (Appendix Table~\ref{tab:eval-hardlabel}).
For the hard label evaluation, personality detection is treated as a classification task.
The model is required to independently classify the user's personality on each dimension, producing outputs such as \textit{CHOICE: A} or \textit{CHOICE: B}.
We present the detailed content of the prompting methods in Appendix  Table~\ref{table:inference_template_hardlabel} for hard labels.
We find that when applying multi-step reasoning methods, the model tends to produce more extreme classification results, as shown in Appendix Figure \ref{fig:mstand}-\ref{fig:mpsy}. This leads to inferior performance of Zero-shot in the hard label metrics compared to the other three methods, while in the soft label evaluation, Zero-shot demonstrates stronger performance.

\section{Related Works}
    The Big Five and MBTI are two widely used personality frameworks in the fields of computational linguistics and natural language processing~\citep{yang2021learning}.
In this paper, we focus on MBTI, one of the most widely used non-clinical psychometric assessments because it translates well into the behavioral context~\citep{tajner2021WhyIM_EACL} and is widely adopted in diverse real-world applications~\citep{kuipers2009influence,Garden2011RelationshipsBM,Gountas2000ANP}.
MBTI distinguishes itself in applied psychological settings by offering an approach that simplifies interpersonal and organizational dynamics, making it particularly valuable for enhancing team functionality and personal development initiatives~\citep{myers1980gifts}.

The quality issues of datasets will affect the accurate evaluation of LLM performance and the iteration of methods in personality detection tasks~\citep{liu-etal-2023-xdailydialog}.
There have been some early explorations of dataset quality in previous work~\citep{tajner2021WhyIM_EACL}, but they only annotated one commonly used Twitter dataset.
These studies do not consider the impact of data noise and do not annotate complete samples. 
Additionally, these works do not validate their conclusions through model experiments.
These limitations indicate that there is still room for improvement in the discussion of dataset quality.
Some studies assess the ability of LLMs to determine the personality of humans or the LLMs themselves~\citep{Ji2023IsCA,caron-srivastava-2023-manipulating,Rao2023CanCA}.

\section{Conclusion}
    
MBTI is one of the most popular personality theories and has garnered considerable research interest over the years.
In this paper, we point out the over-optimism in MBTI personality detection from the nature of population personality traits.
Specifically, (1) the self-reported labels in existing datasets result in data quality issues and (2) the hard labels fail to capture the full range of population personality distributions.
We construct ~\ourdataset, the first high-quality manually annotated MBTI personality detection dataset with soft labels, under the guidance of psychologists.
We filter and annotate the data, and estimate soft labels by deriving the polarity tendency of samples.
Experiment results highlight the polarized predictions and bias as future directions.

\section*{Acknowledgments}

We gratefully acknowledge the support of the National Natural Science Foundation of China (NSFC) via grant 62236004, 62206078, 62441603 and 62476073. 

We are immensely grateful to Professor Mengyue Wu for her insightful guidance and encouragement, which continue to be instrumental in shaping the core contributions of our paper.
We also extend our sincere thanks to Zeming Liu and Jun Xu for their generous contributions of time and expertise in offering suggestions on manuscript writing. Their input significantly enhances the quality and depth of our paper.
We are deeply grateful to Xin Hao for the insightful suggestions regarding the annotation guidelines, which are invaluable to this paper.
We would like to express our profound appreciation to Lina Zhang for her meticulous work on the figures, which have become a highlight of our paper.

\bibliography{colm2024_conference}
\bibliographystyle{colm2024_conference}

\appendix
\label{sec:appendix}
\clearpage

\section{Ethical Statement}
\subsection{Data Access and Privacy}
Our research utilizes data from three sources: PANDORA, Twitter, and Kaggle. 
We obtain special access and adaptation permission for the PANDORA dataset, ensuring compliance with data usage policies. 
The Twitter and Kaggle datasets are publicly available and freely accessible for academic research.
All data used is sourced from these existing datasets, and we do not interact directly with or collect data from social media users.
This ensures that the privacy of individuals is strictly maintained, as we have not accessed or manipulated their social media accounts. 
Once the paper is accepted, we will make our code and dataset publicly available.

\subsection{Annotator Recruitment}
Our three annotators hold undergraduate or master's degrees in English and are fluent in the language, with extensive experience in annotation. Under the guidance of a psychological expert, we provide them with meticulous training, as described in Section~\ref{sec:annotation-training}, enabling them to master the definitions and annotation criteria of MBTI proficiently.
The annotators involved in our study are fairly compensated at rates not lower than the local average wage. They are provided with sufficient time to complete their tasks, ensuring that they can work without experiencing undue fatigue. All annotators give informed consent before participating in the study.

\subsection{Annotation Guidelines and Expert Involvement}Our annotation guidelines are developed under the guidance of psychology experts. Both the pre-annotation and formal annotation recheck processes are conducted with their guidance, ensuring the accuracy and reliability of the annotations.

\subsection{Scope of Analysis}Our analysis focus solely on the textual content of the datasets, specifically examining features related to the MBTI (Myers-Briggs Type Indicator). We do not make any inferences or judgments about the individuals behind the data, such as their habits, gender, opinions, or preferences. Our study strictly concentrates on the text itself and the MBTI-related features it contains, ensuring that no personal attributes or identities are considered in our analysis.

\section{FAQs}
\subsection{The relatively small sample size may limit dataset generalizability?}
(1) Future Scalability: We fully agree that expanding the dataset through semi-automated methods could further enhance its generalizability. In future work, we plan to use LLMs fine-tuned on MBTIBench to label additional data with soft labels, followed by human quality checks.
(2) Current Dataset Scope: In this initial work, our goal is to present \ourdataset as a high-quality, population-aligned dataset that supports MBTI soft-label evaluation. Given the costs of expert-led annotation, we consider the current dataset size to be sufficient for effective model evaluation and to maintain high reliability~\cite{Chen2021EvaluatingLL,Liu2023VisualIT}. We are committed to open-sourcing MBTIBench to facilitate broader use and encourage semi-automated expansion by the community.

\subsection{Potential biases introduced by annotators?}
We recognize the potential impact of annotator bias on consistency.
(1) Annotator Training: We conduct multiple rounds of trial annotations for professional annotators, using both Fleiss' Kappa and accuracy metrics to ensure consistency and reliability in their annotations.
(2) Soft Label Generation: We employ an EM algorithm that accounts for annotator variability and adjusts weights to produce soft labels for each sample (Sec.~\ref{sec:soft-label-generation}). For example, as shown in Appendix Table~\ref{tab:annotator-acc}, annotator A3 has a lower accuracy on P+ labels (63.64\%). The EM algorithm assigns less weight to A3's P+ labels, resulting in a more balanced and realistic distribution, as shown in Appendix Figure~\ref{fig:JP}(a).
We will provide a more detailed explanation in future revisions.

\section{Data Annotation Guidelines}
Personality refers to the combination of characteristics or qualities that form an individual's distinctive character~\citep{Mairesse2007UsingLC}.
It encompasses a wide range of traits, behaviors, thoughts, and emotional patterns that evolve from biological and environmental factors~\citep{Furnham1996TheBF,Celli2018IsBF,chung2011psychological,Line1948DescriptionAM}.
A particular personality can determine various outward observable properties or features, including consistent behavioral patterns, communication style, emotional expression and so on.

To address incorrect labeling issues and accurately label personality types,we refer to the methodology outlined in \citet{tajner2021WhyIM_EACL} to construct our annotation guidelines.
Psychology PhD students participate in the formulation of these guidelines. 
We discuss the personality traits for each dimension based on the dataset, analyzing and adjusting our guidelines through trial annotations.
Finally, we annotate the trial samples, which serve as expert guidelines.
The detailed annotation guidelines are in Table~\ref{app-table:mbti-annotation}.

\begin{sidewaystable*}[hthp]
  \centering   
      \begin{adjustbox}{width=1.0\columnwidth,center}
        \begin{tabular}{>{\centering\arraybackslash}m{9.125em}>{\centering\arraybackslash}m{7.125em}>{\centering\arraybackslash}m{11.625em}>{\centering\arraybackslash}m{21.19em}>{\centering\arraybackslash}m{27.065em}>{\centering\arraybackslash}m{7.625em}}
    \toprule
    \textbf{Dimension} & \textbf{Classification} & \textbf{Linguistic signals} & \textbf{Criteria for Judgment} & \textbf{Examples} & \textbf{Published Articles}  \\
    \midrule
    \multirow{2}[4]{*}{\centering Attitudes} & Extraversion (E)  & Social interaction \newline{}Immediate Response\newline{}More assertive, positive, enthusiastic & Text predominantly features interaction with others or references to social activities;\newline{}Use of intensifiers and exclamation marks;\newline{}We references. & ``hello mate, Greetings to all! "\newline{}``I'm so impressed by your mind reading skills!"\newline{}``You s are so predictable:tongue:" & \multicolumn{1}{c}{\multirow{2}[1]{*}{\begin{tabular}{@{}c@{}}\cite{Pennebaker1999LinguisticSL}\\\cite{tajner2021WhyIM_EACL}\end{tabular}}}\\
\cmidrule{2-5}          & Introversion (I) & Individual activities\newline{}Introspection\newline{}Less assertive & Text predominantly features individual reflection and personal interests;\newline{}Hedging;\newline{}I references. & ``in the dark All my ideals fall apart When they come..."\newline{}``I do have social anxiety."\newline{}``I super anxious all the time and tense." &  \\
    \midrule
    \multirow{2}[4]{*}{\centering Perceiving Functions} & Sensing (S) & Concrete details \newline{}Past/present experiences (reality);\newline{} & Text rich in specific, sensory details and real-life examples;\newline{}Clear, concise, simplified and straightforward writing style. & ``I did a quick google search."\newline{}``I posted some guesses in the anime/manga thread."\newline{}``She called the cops on me and they arrested my whole family for some reason."  & \multicolumn{1}{c}{\multirow{2}[1]{*}{\begin{tabular}{@{}c@{}}~\cite{Mairesse2007UsingLC}\\\cite{Gill2019TakingCO}\end{tabular}}}\\
\cmidrule{2-5}          & iNtuition (N) & Abstract concepts \newline{}Future possibilities & Text rich in abstract ideas and future-oriented thoughts;\newline{}Artistic, longer, complex writing style. & ``I believe strongly in hard work when it is for a good cause, practicality when it..."\newline{}``if money wasn't an issue I'd study forensic psychology and anthropology."\newline{}``Knowing others is intelligence; knowing yourself is true wisdom." &  \\
    \midrule
    \multirow{2}[4]{*}{\centering Judging Functions} & Thinking (T) & Objective\newline{}Logical decision-making & Text displays logical reasoning and analysis;\newline{}Mention of opinions, ideas, comparisons;\newline{}Direct. & ``Poverty is often a result of under education."\newline{}``I think that this is absolutely true. Except for 3 things:  1. In my opinion: There should be one more section in the love-map: Reaction to change."\newline{}``I think the problem is also in the way that we utilize conveniences-Not in the conveniences..." & \multicolumn{1}{c}{\multirow{2}[1]{*}{\begin{tabular}{@{}c@{}}~\cite{Scherer2003VocalCO}\\\cite{Argamon2005LEXICALPO}\\\cite{Kosinski20}\end{tabular}}}\\
\cmidrule{2-5}          &  Feeling (F) & Subjective\newline{}Emotionally-driven choices & Text displays expressions of personal values and emotions;\newline{}Mention of people, values, feelings;\newline{}Tactful, indirect. & ``about to go home - kinda lonely feeling again :/fml.."\newline{}``I like them both :)"\newline{} ``Tired, Half-depressed, and bleh :D" &  \\
    \midrule
    \multirow{2}[4]{*}{\centering Lifestyle Preferences} & Judging (J) & Planning\newline{}Organized & Text shows a preference for planning and organization;\newline{}Past simple tense or present perfect tense;\newline{}Adherence to conventional grammatical standards in a formal writing format. & `` When I'm working out ideas in my head I'll use my whiteboard and gesture to myself while thinking." \newline{}``Right now I am trying to get to 15 posts on this forum."\newline{}``Now that I know I'm going to be a mom, I'm debating whether or not we should vacinate our child."& \multicolumn{1}{c}{\multirow{2}[1]{*}{\begin{tabular}{@{}c@{}}~\cite{Plank2015PersonalityTO}\\\cite{Verhoeven2016TwiStyAM}\\\cite{Yamada2019IncorporatingTI}\end{tabular}}}\\
\cmidrule{2-5}          & Perceiving (P) & Flexibility\newline{}Organized & Text shows openness to new experiences and adaptability;\newline{}Present simple tense;\newline{}Casual writing with occasional lapses in grammar. &  ``For as long as I can remember. As a kid I was in gymnastics, ballet, soccer, piano lessons, you name it. never stuck with any of them."\newline{}``WHAAAAT! That's so insane! Oh my gosh my mind has just been blown."\newline{}``I think!!!l will have tea... :gentleman:" &  \\
    \bottomrule
    \end{tabular}%
    \end{adjustbox}
  \caption{To address incorrect labeling issues and accurately label personality types,we refer to the methodology outlined in \citet{tajner2021WhyIM_EACL} to construct our annotation guidelines.Psychology PhD students participate in the formulation of these guidelines.}
  \label{app-table:mbti-annotation}%
\end{sidewaystable*}

\section{Dataset Quality Analysis}
\subsection{Soft Labels Align with Population Traits}

We illustrate the estimated soft labels and the corresponding sample count distribution of \textit{E/I} in Figure~\ref{fig:IE} and \textit{S/N}, \textit{T/F}, and \textit{J/P} in Appendix Figure~\ref{fig:SN}-\ref{fig:JP}.
Soft labels exhibit a smooth distribution from 0 to 1, and there are more non-extreme samples than extreme ones.
This is consistent with the population distribution and demonstrates that our soft labels align with population traits better compared to hard labels~\citep{Harvey1994ScoringTM}.

\begin{figure*}[!ht]
	\centering
    \vspace{0.4cm}
	\includegraphics[width=1.0\linewidth]{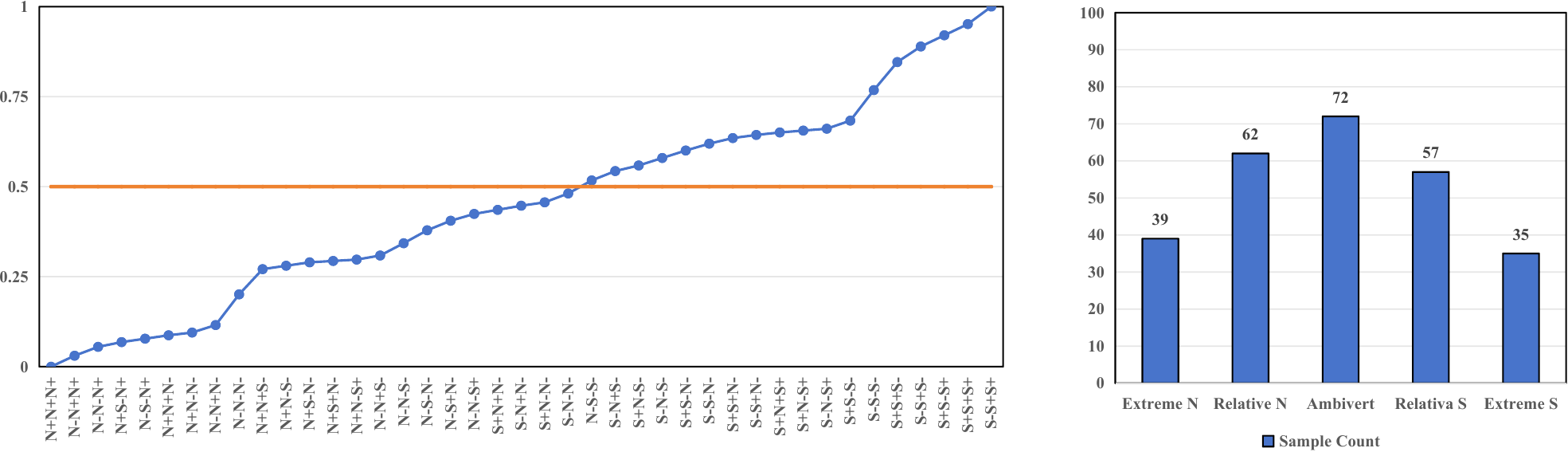}
    \caption{
    Distribution and Polarity distribution of soft labels for the \textit{S/N} dimension in
\ourdataset.
    }
\label{fig:SN}
\end{figure*}

\begin{figure*}[!ht]
	\centering
    \vspace{0.4cm}
	\includegraphics[width=1.0\linewidth]{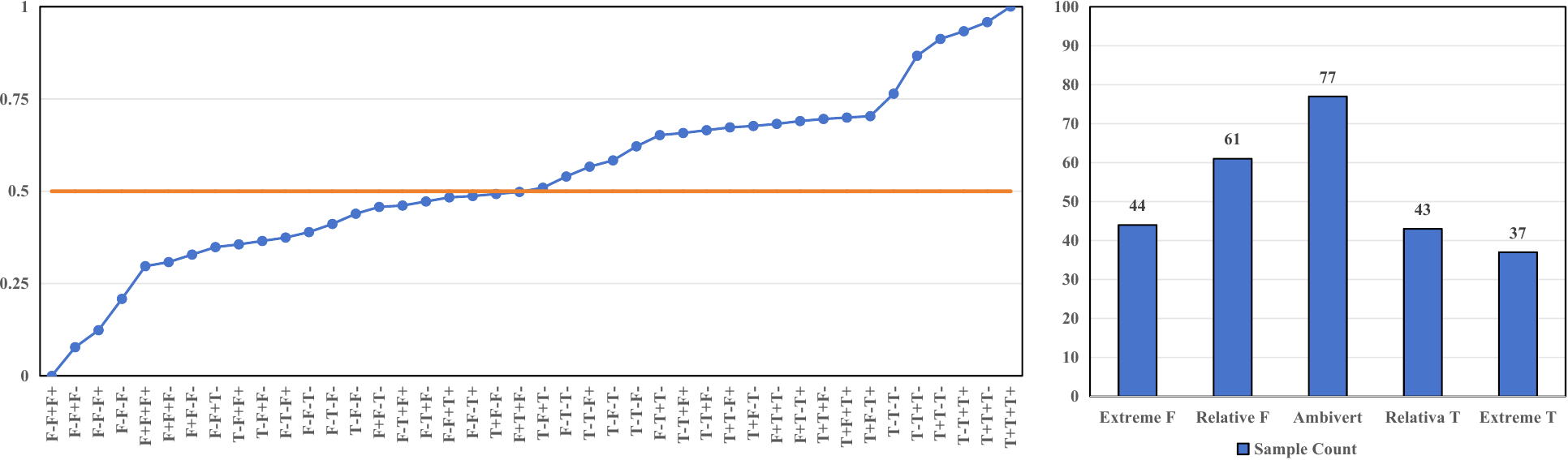}
    \caption{
    Distribution and Polarity distribution of soft labels for the \textit{T/F} dimension in
\ourdataset.
    }
\label{fig:TF}
\end{figure*}

\begin{figure*}[!ht]
	\centering
    \vspace{0.4cm}
	\includegraphics[width=1.0\linewidth]{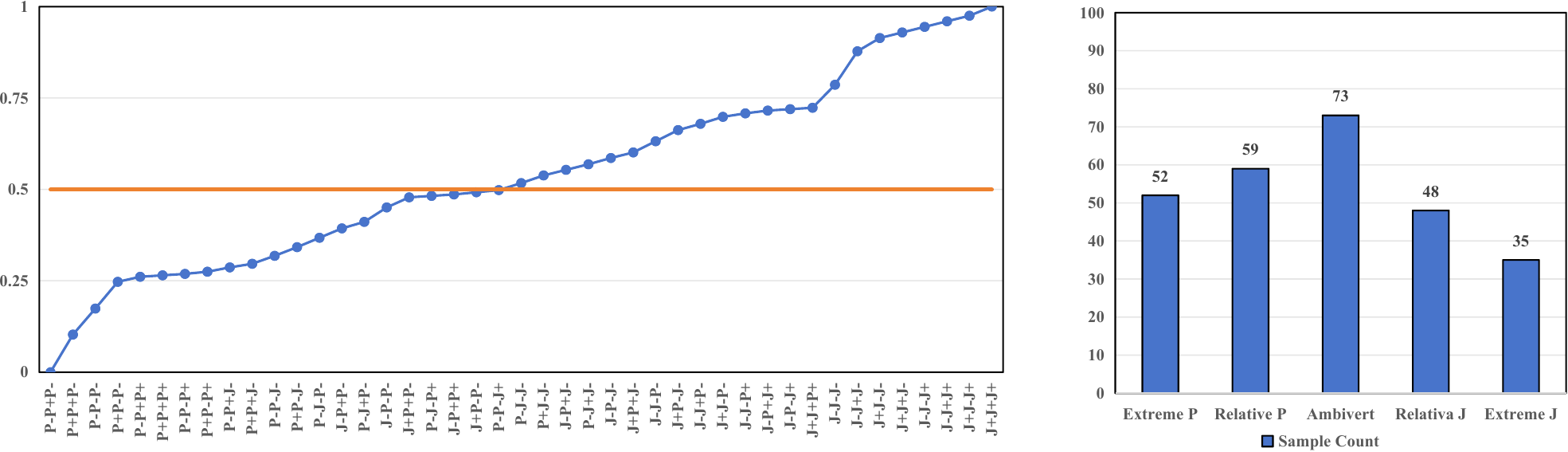}
    \caption{
    Distribution and Polarity distribution of soft labels for the \textit{J/P} dimension in
\ourdataset
    }
\label{fig:JP}
\end{figure*}

\section{Experimental Setup}
\label{app-sec:Experimental_Setup}

\begin{table*}[!ht]
\centering
\small
\begin{adjustbox}{width=1.0\textwidth}
\begin{tabular}{p{3cm}|p{12cm}}
\toprule
\textbf{Method} & \textbf{Template} \\ 
\midrule
Zero-shot & System: Given the following text from a user's social media posts, determine the first dimension (Extraversion or Introversion) of Myers-Briggs Type Indicator (MBTI) personality type best fits the user. You need to rate the statement with a score 0-10, where 0=more E and 10=more I, output your final score by strictly following this format: ``[[score]]'' and do not give reason. \newline User: \textit{\textless ALL POSTS \textgreater} \newline Assistant: [[score]] \\
\midrule
Step-by-step & System: Given the following text from a user's social media posts, determine the first dimension (Extraversion or Introversion) of Myers-Briggs Type Indicator (MBTI) personality type best fits the user. You need to rate the statement with a score 0-10, where 0=more E and 10=more I, output your final score by strictly following this format: ``[[score]]''. Let's think step by step. \newline User: \textit{\textless ALL POSTS \textgreater} \newline Assistant: \textit{\textless Thinking step-by-step \textgreater} \newline User: According to above, what is the score of EI dimension. Output your final score by strictly following this format: ``[[score]]'' and do not give reason.\newline Assistant: CHOICE: [[score]] \\
\midrule
Few-shot & System: Given the following text from a user's social media posts, determine the first dimension (Extraversion or Introversion) of Myers-Briggs Type Indicator (MBTI) personality type best fits the user. You need to rate the statement with a score 0-10, where 0=more E and 10=more I, output your final score by strictly following this format: ``[[score]]''  and do not give reason. \newline User: Consider the first example: \textit{\textless First example \textgreater} \newline The score is [[score]] \newline Consider the second example: \textit{\textless Second example \textgreater} \newline The score is [[score]] \newline Consider the third example: \textit{\textless ALL POSTS \textgreater} \newline The score is \newline Assistant: [[score]] \\
\midrule
PsyCoT & System: You are an AI assistant who specializes in text analysis and I am User. We will complete a text analysis task together through a multi-turn dialogue. The task is as follows: we have a set of posts written by an author, and at each turn I will give you a Question about the author. According to the author's posts, you need to choose the possible options ONLY. DO NOT give your reason, just wait for the next user input. After opting all the choices, I will ask you the EI dimension (Extraversion or Introversion) score of the author. You need to rate the statement with a score 0-10, where 0=more E and 10=more I. \newline AUTHOR'S POSTS: \textit{\textless ALL POSTS \textgreater} \newline User: Q: The author is usually: A: "A good mixer with gropus of people", B: "Quiet and reserved", or C: "Not sure whether A or B". Provide a choice ID in the format: "CHOICE: \textless A/B/C \textgreater" only, and do not give the explanation. do not generate User input. \newline Assistant: CHOICE: \textit{\textless A or B \textgreater} \newline $\ldots$ ( \textit{\textless Questionnaires \textgreater} ) \newline User: According to above, what is the score of EI dimension. Output your final score by strictly following this format: ``[[score]]'' and do not give reason. \newline Assistant: [[score]] \\
\bottomrule
\end{tabular}
\end{adjustbox}
\caption{Inference prompt templates for soft labels.}
\label{table:inference_template_softlabel}
\end{table*}

\begin{table*}[!ht]
\centering
\small
\begin{adjustbox}{width=1.0\textwidth}
\begin{tabular}{p{3cm}|p{12cm}}
\toprule
\textbf{Method} & \textbf{Template} \\ 
\midrule
Zero-shot & System: Given the following text from a user's social media posts, determine the first dimension of Myers-Briggs Type Indicator (MBTI) personality type best fits the user. Predicting whether the author is A: ``Introversion'' or B: ``Extraversion''. Provide a choice in the format: `CHOICE: \textless{}A/B\textgreater{}' and do not give reason \newline User: \textit{\textless ALL POSTS \textgreater} \newline Assistant: CHOICE: \textit{\textless A or B \textgreater} \\
\midrule
Step-by-step & System: Given the following text from a user's social media posts, determine the first dimension of Myers-Briggs Type Indicator (MBTI) personality type best fits the user. Predicting whether the author is A: ``Introversion'' or B: ``Extraversion''. Let's think step by step. Finally a choice in the format: `CHOICE: \textless{}A/B\textgreater{}' and do not give reason \newline User: \textit{\textless ALL POSTS \textgreater} \newline Assistant: \textit{\textless Thinking step-by-step \textgreater} \newline User: According to above, the author is more likely to be: A: "Introversion" or B: "Extraversion". Provide a choice in the format: "CHOICE: \textless A/B \textgreater" and do not give the explanation.\newline Assistant: CHOICE: \textit{\textless A or B \textgreater} \\
\midrule
Few-shot & System: Given the following text from a user's social media posts, determine the first dimension of Myers-Briggs Type Indicator (MBTI) personality type best fits the user. Predicting whether the author is A: ``Introversion'' or B: ``Extraversion''. Provide a choice in the format: `CHOICE: \textless{}A/B\textgreater{}' and do not give reason \newline User: Consider the first example: \textit{\textless First example \textgreater} \newline The choice is CHOICE: \textit{\textless A or B \textgreater} \newline Consider the second example: \textit{\textless Second example \textgreater} \newline The choice is CHOICE: \textit{\textless A or B \textgreater} \newline Consider the third example: \textit{\textless ALL POSTS \textgreater} \newline The choice is \newline Assistant: CHOICE: \textit{\textless A or B \textgreater} \\
\midrule
PsyCoT & System: You are an AI assistant who specializes in text analysis and I am User. We will complete a text analysis task together through a multi-turn dialogue. The task is as follows: we have a set of posts written by an author, and at each turn I will give you a Question about the author. According to the author's posts, you need to choose the possible options ONLY. DO NOT give your reason, just wait for the next user input. After opting all the choices, I will ask you if the author is A: "Introversion" or B: "Extraversion", and then you need to give your choice. \newline AUTHOR'S POSTS: \textit{\textless ALL POSTS \textgreater} \newline User: Q: The author is usually: A: "A good mixer with gropus of people", B: "Quiet and reserved", or C: "Not sure whether A or B". Provide a choice ID in the format: "CHOICE: \textless A/B/C \textgreater" only, and do not give the explanation. do not generate User input. \newline Assistant: CHOICE: \textit{\textless A or B \textgreater} \newline $\ldots$ ( \textit{\textless Questionnaires \textgreater} ) \newline User: According to above, the author is more likely to be: A: "Introversion" or B: "Extraversion". Provide a choice in the format: "CHOICE: \textless A/B \textgreater" and do not give the explanation. \newline Assistant: CHOICE: \textit{\textless A or B \textgreater} \\
\bottomrule
\end{tabular}
\end{adjustbox}
\caption{Inference prompt templates for hard labels.}
\label{table:inference_template_hardlabel}
\end{table*}

\subsection{Prompting Methods}
Following the methodologies outlined in \citet{Yang2023PsyCoTPQ}, based on each user's posts $P$, we apply them to an inference prompt template $t(\cdot)$, where $t(P)$ serves as the actual input to LLMs.
We adopt four prompting approaches including Zero-shot, Step-by-step~\citep{kojima2022large}, Few-shot, and PsyCoT~\citep{Yang2023PsyCoTPQ}.
We present the detailed content of the prompting methods of soft labels in Appendix Table~\ref{table:inference_template_softlabel}.

\subsection{Metrics}
\label{app-sec:metrics}
\begin{figure}[!ht]
	\centering
    \vspace{0.4cm}
	\includegraphics[width=0.7\linewidth]{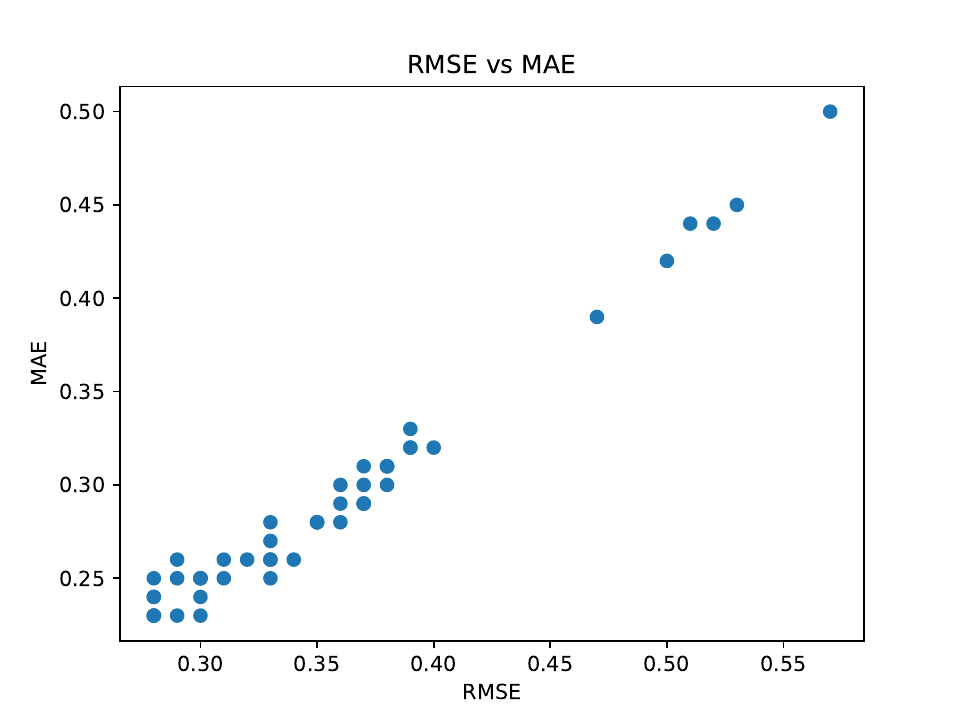}
    \caption{
Consistency between RMSE and MAE.
    }
\label{fig:rmse_mae}
\end{figure}

We use accuracy (Acc) and macro F1 score as evaluation metrics for hard labels. For soft labels, we treat human-annotated labels and model outputs as two continuous distributions. We employ the following metrics to measure the differences between them:\footnote{We also analyze the consistency between MAE and RMSE as soft label metrics, as illustrated in Figure~\ref{fig:rmse_mae}. RMSE and MAE exhibit a strong linear correlation. 
}

\subsubsection{Segmented Root Mean Square Error}
We adopt a similar approach to ~\citeauthor{raykar2010learning}, using the Root Mean Square Error (RMSE) to measure the average distance between the predicted and actual values of the model.
Since we map the continuous 0-1 distribution into 10 discrete intervals, the S-RMSE calculation is performed using the indices of these intervals.
\begin{equation}
    \text{S-RMSE} = \sqrt{\frac{1}{n} \sum_{i=1}^n (bin(y_i) - bin(\hat{y}_i))^2}
\end{equation}
Here, $y_i$ represents the actual values, $\hat{y}_i$ denotes the predicted values, $bin(\cdot)$ maps the continuous distribution into discrete intervals, and $n$ is the total number of samples.
Due to the squaring operation, S-RMSE is particularly sensitive to large errors.
\subsubsection{Segmented Mean Absolute Error}

We adopt a method similar to ~\cite{Willmott2005AdvantagesOT}, using the Mean Absolute Error (MAE) to measure the average difference between the predicted and actual values. Similar to S-RMSE, we adopt S-MAE on discrete intervals.
\begin{equation}
\text{S-MAE} = \frac{1}{n} \sum_{i=1}^{n} |bin(y_i) - bin(\hat{y}_i)|
\end{equation}
where $y_i$ represents the actual value, $\hat{y}_i$ represents the predicted value, $bin(\cdot)$ maps the continuous distribution into discrete intervals, and $n$ is the total number of samples. Compared to the Segmented Root Mean Square Error (S-RMSE), S-MAE is less sensitive to outliers as it calculates the error using absolute differences. This makes S-MAE provide a more straightforward interpretation of average model error, making it suitable for scenarios where the goal is to understand the typical magnitude of prediction errors without being influenced by extreme values.

\subsection{Experimental Details for Hard Label Prediction}
We directly match the model outputs to the options \textit{CHOICE: A} and \textit{CHOICE: B}. 
Take \textit{E/I} as an example, considering that some models have limited instruction-following capabilities, we also accept outputs like \textit{CHOICE: \textless A\textgreater}, \textit{CHOICE: a}, \textit{CHOICE: E}, and \textit{CHOICE: Introversion}. 
Although these outputs do not strictly match the correct answer, they reveal the model's preference for a particular personality option.
We attribute this to a deficiency in the model's ability to follow instructions rather than a flaw in its personality detection capabilities. 
Therefore, we accept these outputs as well.
If the model's output does not match any of the specified rules, such as \textit{CHOICE: C} or \textit{CHOICE: Cannot give answers}, we categorize it as an invalid category. We present the detailed content of the prompting methods of hard labels in Appendix Table~\ref{table:inference_template_hardlabel}.

\section{Results and Analysis}

\subsection{Soft Label Evaluation with MAE and RMSE}

In this section, we introduce the model evaluation results with MAE and RMSE.
The results are in Appendix Table~\ref{tab:eval_softlabel_raw}. 
Zero-shot is the best method on most backbones (5 out of 6).
Considering the limited precision of our estimated soft labels, %we introduce Segmented MAE (S-MAE) and Segmented RMSE (S-RMSE).
the errors tend to be sensitive to specific small deviations in continuous values when applying MAE and RMSE.
To address this issue, we introduce Segmented MAE (S-MAE) and Segmented RMSE (S-RMSE). 
Instead of computing the error based solely on the raw predictions, we discretize the continuous 0-1 distribution into nine intervals and evaluate the corresponding RMSE and MAE within these segments.
Therefore, the results with S-MAE and S-RMSE in Sec.~\ref{sec:model-evaluation} focus more on whether the prediction falls within an appropriate range.

\setlength{\tabcolsep}{1pt} % Adjust this value to reduce column spacing
\begin{table*}[!ht]
  \centering
\begin{adjustbox}{width=0.95\textwidth}
    % \small
    \footnotesize

\begin{tabular}{c@{\hspace{5pt}}c@{\hspace{8pt}}c@{\hspace{5pt}}c@{\hspace{8pt}}c@{\hspace{5pt}}c@{\hspace{8pt}}c@{\hspace{5pt}}c@{\hspace{8pt}}c@{\hspace{5pt}}c@{\hspace{8pt}}c}
    \toprule
    \multirow{2}[2]{*}{\textbf{Backbones}} & \multirow{2}[2]{*}{\textbf{Methods}} & \multicolumn{2}{c}{\textbf{E/I}} & \multicolumn{2}{c}{\textbf{S/N}} & \multicolumn{2}{c}{\textbf{T/F}} & \multicolumn{2}{c}{\textbf{J/P}} & \multirow{2}[2]{*}{\textbf{Rank}}                                                                                 \\
                                           &                                      & \textbf{RMSE}                    & \textbf{MAE}                     & \textbf{RMSE}                    & \textbf{MAE}                     & \textbf{RMSE}                     & \textbf{MAE}      & \textbf{RMSE}     & \textbf{MAE}      &                   \\
    \midrule
    \multirow{4}[2]{*}{gpt-4o-mini}        & Zero-shot                            & \textbf{0.320}                   & \textbf{0.240}                   & \textbf{0.310}                   & \textbf{0.250}                   & \textbf{0.270}                    & \textbf{0.220}    & \textbf{0.340}    & \textbf{0.280}    & \textbf{1.000}    \\
                                           & Step-by-step                         & \underline{0.330}                & \underline{0.250}                & 0.350                            & 0.280                            & 0.320                             & 0.250             & 0.400             & 0.320             & 2.750             \\
                                           & Few-shot                             & \underline{0.330}                & 0.270                            & \underline{0.320}                & \underline{0.270}                & \underline{0.290}                 & \underline{0.240} & \textbf{0.340}    & \textbf{0.280}    & \underline{1.875} \\
                                           & PsyCoT                               & 0.350                            & 0.270                            & 0.510                            & 0.420                            & 0.400                             & 0.330             & 0.460             & 0.360             & 3.875             \\
    \midrule
    \multirow{4}[2]{*}{gpt-4o}             & Zero-shot                            & \underline{0.330}                & \underline{0.260}                & \textbf{0.330}                   & \textbf{0.260}                   & \textbf{0.300}                    & \textbf{0.240}    & 0.420             & 0.350             & \underline{1.750} \\
                                           & Step-by-step                         & 0.350                            & 0.270                            & \underline{0.340}                & \underline{0.270}                & \underline{0.310}                 & 0.250             & \textbf{0.390}    & \underline{0.320} & 2.250             \\
                                           & Few-shot                             & \textbf{0.320}                   & \textbf{0.240}                   & 0.380                            & 0.310                            & \underline{0.310}                 & \textbf{0.240}    & \textbf{0.390}    & \textbf{0.310}    & \textbf{1.625}    \\
                                           & PsyCoT                               & 0.360                            & 0.290                            & 0.520                            & 0.440                            & 0.350                             & 0.270             & 0.530             & 0.430             & 4.000             \\
    \midrule
    \multirow{4}[2]{*}{Qwen2-7B}           & Zero-shot                            & \underline{0.350}                & \underline{0.290}                & \textbf{0.300}                   & \textbf{0.260}                   & \underline{0.320}                 & \textbf{0.270}    & \underline{0.340} & \underline{0.290} & \underline{1.625} \\
                                           & Step-by-step                         & 0.370                            & 0.300                            & \textbf{0.300}                   & \textbf{0.260}                   & \underline{0.320}                 & 0.280             & \underline{0.340} & \underline{0.290} & 2.125             \\
                                           & Few-shot                             & 0.380                            & 0.310                            & 0.360                            & 0.300                            & 0.450                             & 0.360             & \underline{0.340} & \underline{0.290} & 3.500             \\
                                           & PsyCoT                               & \textbf{0.290}                   & \textbf{0.260}                   & \textbf{0.300}                   & \textbf{0.260}                   & \textbf{0.310}                    & \textbf{0.270}    & \textbf{0.330}    & \textbf{0.280}    & \textbf{1.000}    \\
    \midrule
    \multirow{4}[2]{*}{Qwen2-72B}          & Zero-shot                            & \textbf{0.280}                   & \underline{0.240}                & \textbf{0.300}                   & \textbf{0.260}                   & \underline{0.330}                 & \underline{0.270} & \textbf{0.330}    & \textbf{0.280}    & \underline{1.375} \\
                                           & Step-by-step                         & \textbf{0.280}                   & \textbf{0.230}                   & \textbf{0.300}                   & \textbf{0.260}                   & \textbf{0.320}                    & \textbf{0.260}    & \textbf{0.330}    & \textbf{0.280}    & \textbf{1.000}    \\
                                           & Few-shot                             & 0.310                            & 0.250                            & 0.350                            & 0.290                            & 0.370                             & 0.310             & \textbf{0.330}    & \textbf{0.280}    & 2.750             \\
                                           & PsyCoT                               & 0.290                            & \underline{0.240}                & 0.460                            & 0.370                            & 0.510                             & 0.420             & 0.530             & 0.440             & 3.625             \\
    \midrule
    \multirow{4}[2]{*}{Llama3.1-8B}        & Zero-shot                            & \textbf{0.310}                   & \textbf{0.270}                   & \textbf{0.320}                   & \textbf{0.270}                   & \textbf{0.310}                    & \textbf{0.270}    & \textbf{0.330}    & \textbf{0.290}    & \textbf{1.000}    \\
                                           & Step-by-step                         & 0.430                            & 0.360                            & 0.440                            & 0.360                            & 0.450                             & 0.370             & 0.410             & 0.330             & 3.750             \\
                                           & Few-shot                             & 0.410                            & 0.340                            & \underline{0.390}                & \underline{0.310}                & \underline{0.410}                 & \underline{0.340} & \underline{0.380} & \underline{0.320} & \underline{2.250} \\
                                           & PsyCoT                               & \underline{0.380}                & \underline{0.320}                & 0.420                            & 0.350                            & 0.420                             & 0.350             & 0.410             & 0.350             & 2.875             \\
    \midrule
    \multirow{4}[2]{*}{Llama3.1-70B}       & Zero-shot                            & \textbf{0.300}                   & \textbf{0.250}                   & \textbf{0.360}                   & \textbf{0.300}                   & \underline{0.340}                 & \underline{0.280} & \textbf{0.360}    & \textbf{0.300}    & \textbf{1.250}    \\
                                           & Step-by-step                         & \underline{0.330}                & \underline{0.270}                & \underline{0.370}                & \textbf{0.300}                   & \underline{0.340}                 & \underline{0.280} & 0.370             & 0.310             & 2.125             \\
                                           & Few-shot                             & \underline{0.330}                & \underline{0.270}                & \underline{0.370}                & \textbf{0.300}                   & \textbf{0.320}                    & \textbf{0.260}    & \textbf{0.360}    & \textbf{0.300}    & \underline{1.375} \\
                                           & PsyCoT                               & 0.340                            & \underline{0.270}                & 0.430                            & 0.350                            & 0.410                             & 0.330             & 0.430             & 0.350             & 3.750             \\
    \bottomrule
\end{tabular}%
  \end{adjustbox}
  \caption{We conduct experiments using soft labels and use RMSE and MAE as evaluation metrics, where lower values indicate better performance. The best and second-best results across the four methods for each backbone are \textbf{bolded} and \underline{underlined}.}
  % \vspace{-0.4cm}
  \label{tab:eval_softlabel_raw}%
\end{table*}%

\subsection{Hard Label Evaluation}
\label{sec-app:Hard_Label_Evaluation}
\begin{table*}[!ht]
  \centering
  \small
  \setlength{\tabcolsep}{3pt}
    \begin{adjustbox}{width=0.95\textwidth}
        % \small
        \footnotesize
        % \scriptsize
        \begin{tabular}{cccccccccccc}
    \toprule
    \multirow{2}[2]{*}{\textbf{Backbones}} & \multirow{2}[2]{*}{\textbf{Methods}} & \multicolumn{2}{c}{\textbf{E/I}} & \multicolumn{2}{c}{\textbf{S/N}} & \multicolumn{2}{c}{\textbf{T/F}} & \multicolumn{2}{c}{\textbf{J/P}} & \multicolumn{2}{c}{\textbf{Overall}}                                                                                                               \\
                                           &                                      & \textbf{Acc}                     & \textbf{F1}                      & \textbf{Acc}                     & \textbf{F1}                      & \textbf{Acc}                         & \textbf{F1}         & \textbf{Acc}        & \textbf{F1}         & \textbf{Acc}        & \textbf{F1}         \\
    \midrule
    \multirow{4}[2]{*}{gpt-4o-mini}        & Zero-shot                            & 64.69                            & \underline{ 64.02 }              & \underline{ 54.90 }              & \underline{ 53.29 }              & 60.49                                & 54.33               & \underline{ 58.39 } & \underline{ 55.82 } & 59.62               & 56.87               \\
                                           & Step-by-step                         & \textbf{ 66.43 }                 & 63.88                            & \textbf{ 59.44 }                 & \textbf{ 59.42 }                 & \underline{ 72.38 }                  & \underline{ 71.68 } & \textbf{ 59.09 }    & 54.92               & \textbf{ 64.34 }    & \textbf{ 62.48 }    \\
                                           & Few-shot                             & 61.19                            & 61.08                            & 53.50                            & 34.79                            & 66.78                                & 65.86               & 58.04               & 51.36               & 59.88               & 53.27               \\
                                           & PsyCoT                               & \underline{ 66.08 }              & \textbf{ 65.96 }                 & 52.45                            & 45.20                            & \textbf{ 74.13 }                     & \textbf{ 73.71 }    & 57.69               & \textbf{ 56.62 }    & \underline{ 62.59 } & \underline{ 60.37 } \\
    \midrule
    \multirow{4}[2]{*}{gpt-4o}             & Zero-shot                            & \underline{ 66.08 }              & \underline{ 66.06 }              & \underline{ 61.19 }              & \underline{ 60.02 }              & 70.28                                & 68.96               & 57.69               & 48.83               & \underline{ 63.81 } & \underline{ 60.97 } \\
                                           & Step-by-step                         & \textbf{ 67.83 }                 & \textbf{ 66.98 }                 & \textbf{ 65.03 }                 & \textbf{ 64.26 }                 & \textbf{ 72.73 }                     & \textbf{ 72.29 }    & \textbf{ 61.19 }    & \textbf{ 59.46 }    & \textbf{ 66.70 }    & \textbf{ 65.75 }    \\
                                           & Few-shot                             & 61.19                            & 40.69                            & 55.94                            & 36.10                            & \underline{ 72.03 }                  & 47.68               & \underline{ 59.79 } & \underline{ 55.26 } & 62.24               & 44.93               \\
                                           & PsyCoT                               & 62.24                            & 42.40                            & 48.95                            & 29.28                            & 69.58                                & \underline{ 69.00 } & 53.85               & 40.36               & 58.66               & 45.26               \\
    \midrule
    \multirow{4}[2]{*}{Qwen2-7B}           & Zero-shot                            & 59.79                            & 59.48                            & \underline{ 51.05 }              & 45.93                            & 52.80                                & 44.78               & 52.45               & 34.40               & 54.02               & 46.15               \\
                                           & Step-by-step                         & \underline{ 61.89 }              & \underline{ 61.81 }              & 49.65                            & 45.14                            & 52.80                                & 41.65               & \underline{ 52.80 } & 36.47               & \underline{ 54.29 } & 46.27               \\
                                           & Few-shot                             & 46.85                            & 41.84                            & \underline{ 51.05 }              & \underline{ 49.10 }              & \underline{ 55.94 }                  & \underline{ 50.61 } & \textbf{ 53.15 }    & \textbf{ 52.76 }    & 51.75               & \underline{ 48.58 } \\
                                           & PsyCoT                               & \textbf{ 63.29 }                 & \textbf{ 63.00 }                 & \textbf{ 55.24 }                 & \textbf{ 55.10 }                 & \textbf{ 68.88 }                     & \textbf{ 68.77 }    & 47.55               & \underline{ 46.03 } & \textbf{ 58.74 }    & \textbf{ 58.23 }    \\
    \midrule
    \multirow{4}[2]{*}{Qwen2-72B}          & Zero-shot                            & \underline{ 63.64 }              & \underline{ 58.10 }              & 53.15                            & \textbf{ 50.84 }                 & \textbf{ 69.58 }                     & \textbf{ 69.58 }    & 56.64               & 51.88               & \underline{ 60.75 } & \textbf{ 57.60 }    \\
                                           & Step-by-step                         & \textbf{ 66.43 }                 & \textbf{ 62.92 }                 & \textbf{ 63.64 }                 & \underline{ 41.92 }              & \underline{ 68.53 }                  & \underline{ 45.57 } & \underline{ 56.99 } & \underline{ 52.61 } & \textbf{ 63.90 }    & \underline{ 50.76 } \\
                                           & Few-shot                             & 55.59                            & 35.11                            & \underline{ 55.94 }              & 37.92                            & 65.73                                & 43.96               & 56.64               & 31.50               & 58.48               & 37.12               \\
                                           & PsyCoT                               & 44.76                            & 32.82                            & 50.00                            & 30.24                            & 41.61                                & 35.36               & \textbf{ 59.79 }    & \textbf{ 59.18 }    & 49.04               & 39.40               \\
    \midrule
    \multirow{4}[2]{*}{Llama3.1-8B}        & Zero-shot                            & 51.40                            & 51.02                            & 53.85                            & 35.78                            & 52.80                                & 30.40               & 52.45               & 35.06               & 52.63               & 38.07               \\
                                           & Step-by-step                         & \textbf{ 60.14 }                 & \textbf{ 59.86 }                 & \textbf{ 61.54 }                 & \textbf{ 61.52 }                 & \textbf{ 65.73 }                     & \textbf{ 65.45 }    & \textbf{ 56.29 }    & \textbf{ 50.87 }    & \textbf{ 60.93 }    & \textbf{ 59.43 }    \\
                                           & Few-shot                             & \underline{ 54.90 }              & \underline{ 54.36 }              & \underline{ 54.20 }              & \underline{ 53.72 }              & \underline{ 61.19 }                  & \underline{ 60.53 } & \underline{ 55.24 } & \underline{ 46.06 } & \underline{ 56.38 } & \underline{ 53.67 } \\
                                           & PsyCoT                               & 46.15                            & 37.20                            & 51.40                            & 42.78                            & 57.34                                & 48.59               & 53.15               & 36.01               & 52.01               & 41.15               \\
    \midrule
    \multirow{4}[2]{*}{Llama3.1-70B}       & Zero-shot                            & \underline{ 52.80 }              & \textbf{ 37.59 }                 & 46.50                            & 28.59                            & 63.64                                & \underline{ 45.57 } & 52.10               & 28.45               & 53.76               & 35.05               \\
                                           & Step-by-step                         & \textbf{ 55.24 }                 & \underline{ 37.41 }              & \textbf{ 52.80 }                 & \textbf{ 33.96 }                 & \underline{ 67.83 }                  & 44.64               & \underline{ 54.90 } & \underline{ 31.23 } & \textbf{ 57.69 }    & \underline{ 36.81 } \\
                                           & Few-shot                             & 51.05                            & 35.57                            & \underline{ 48.25 }              & \underline{ 30.03 }              & \textbf{ 68.53 }                     & \textbf{ 46.37 }    & \textbf{ 60.84 }    & \textbf{ 39.85 }    & \underline{ 57.17 } & \textbf{ 37.96 }    \\
                                           & PsyCoT                               & 50.00                            & 36.55                            & 33.92                            & 22.54                            & 57.34                                & 42.62               & 34.97               & 28.74               & 44.06               & 32.61               \\
    \bottomrule
\end{tabular}%
    \end{adjustbox}
      \caption{We first perform classification experiments on our dataset using hard labels.}
        \vspace{-0.4cm}  \label{tab:eval-hardlabel}%
\end{table*}%

We provide the results of six LLMs with hard labels in Table~\ref{tab:eval-hardlabel}.

\begin{figure*}[!t]
	\centering
    \vspace{0.4cm}
    \includegraphics[width=1.0\linewidth]{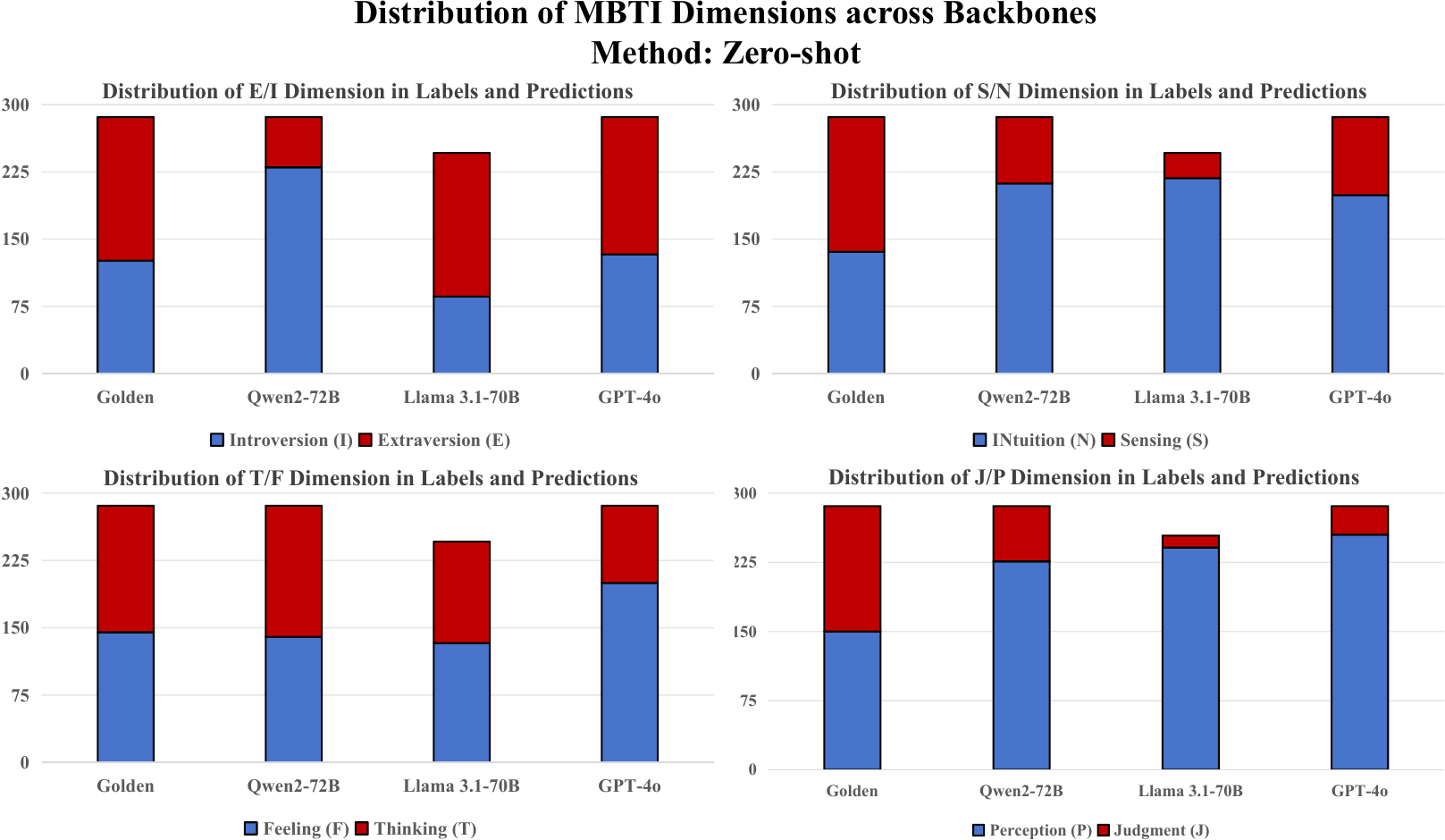}
    \caption{
Model prediction bias on \textit{Zero-shot}.
    }
\label{fig:mstand}
\end{figure*}

\begin{figure*}[!t]
	\centering
    \vspace{0.4cm}
	\includegraphics[width=1.0\linewidth]{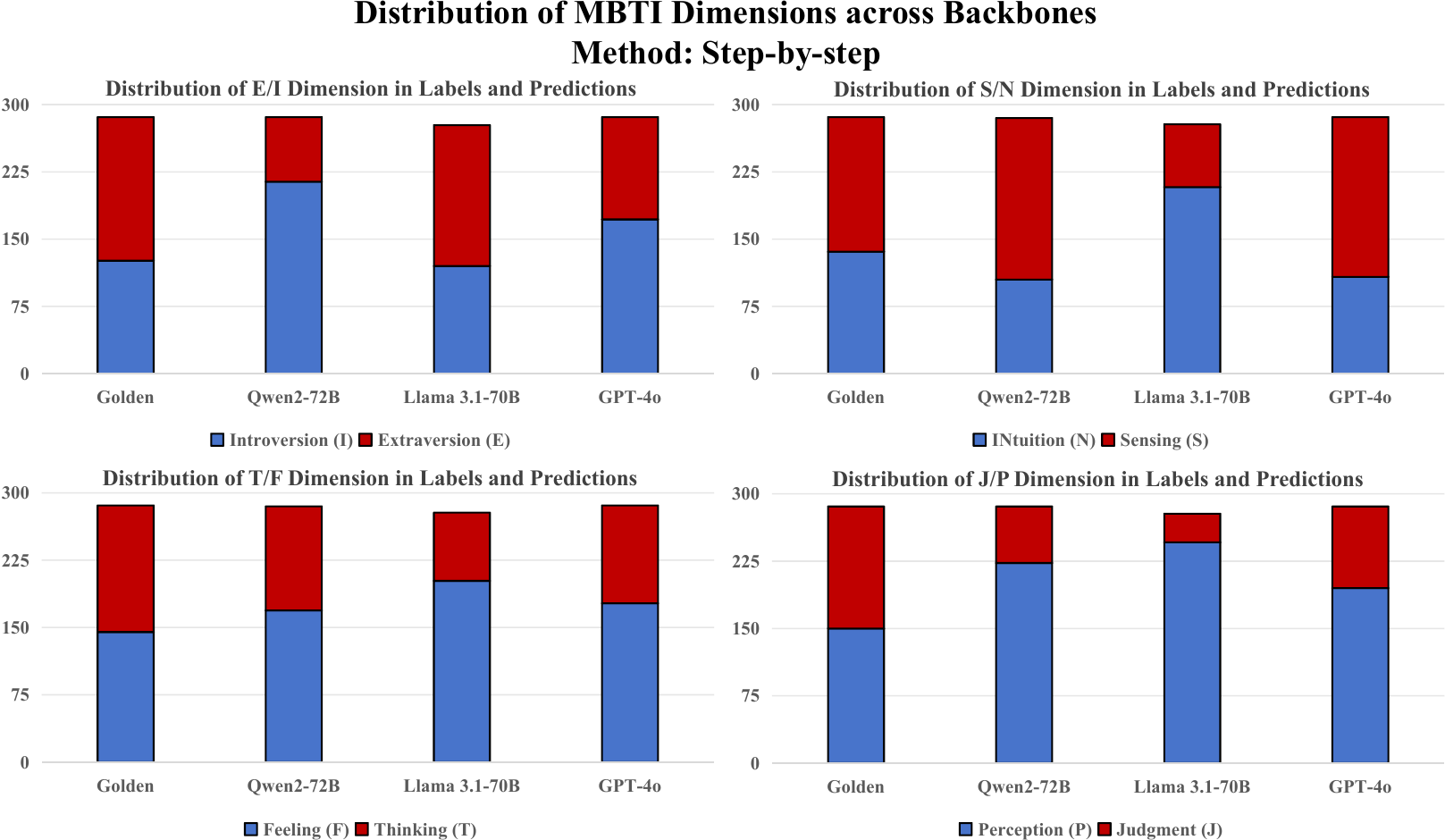}
    \caption{
Model prediction bias on \textit{Step-by-step}.
    }
\label{fig:mstep}
\end{figure*}

\begin{figure*}[!t]
	\centering
    \vspace{0.4cm}
	\includegraphics[width=1.0\linewidth]{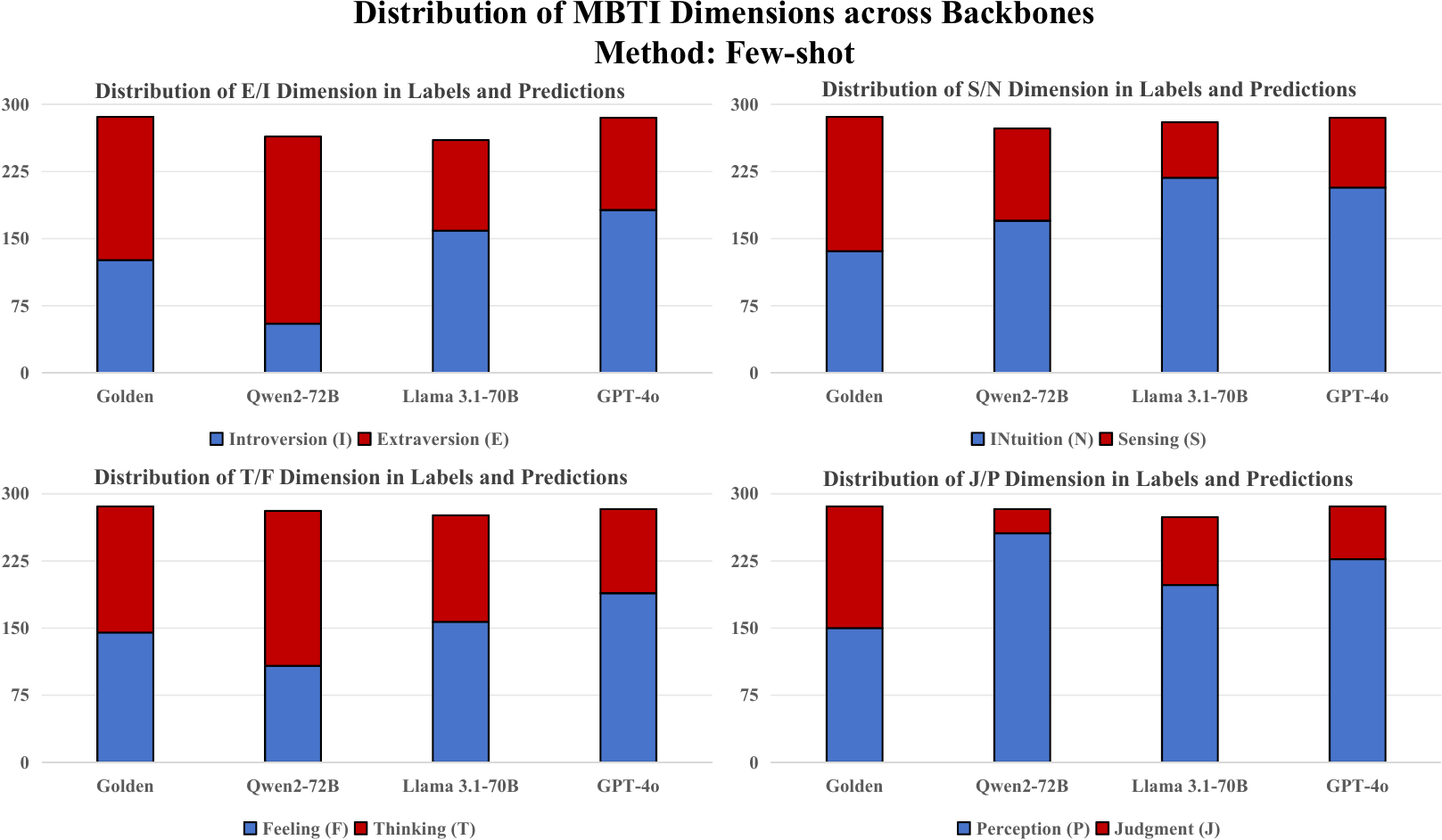}
    \caption{
Model prediction bias on \textit{Few-shot}.
    }
\label{fig:mfew}
\end{figure*}

\begin{figure*}[!t]
	\centering
    \vspace{0.4cm}
	\includegraphics[width=1.0\linewidth]{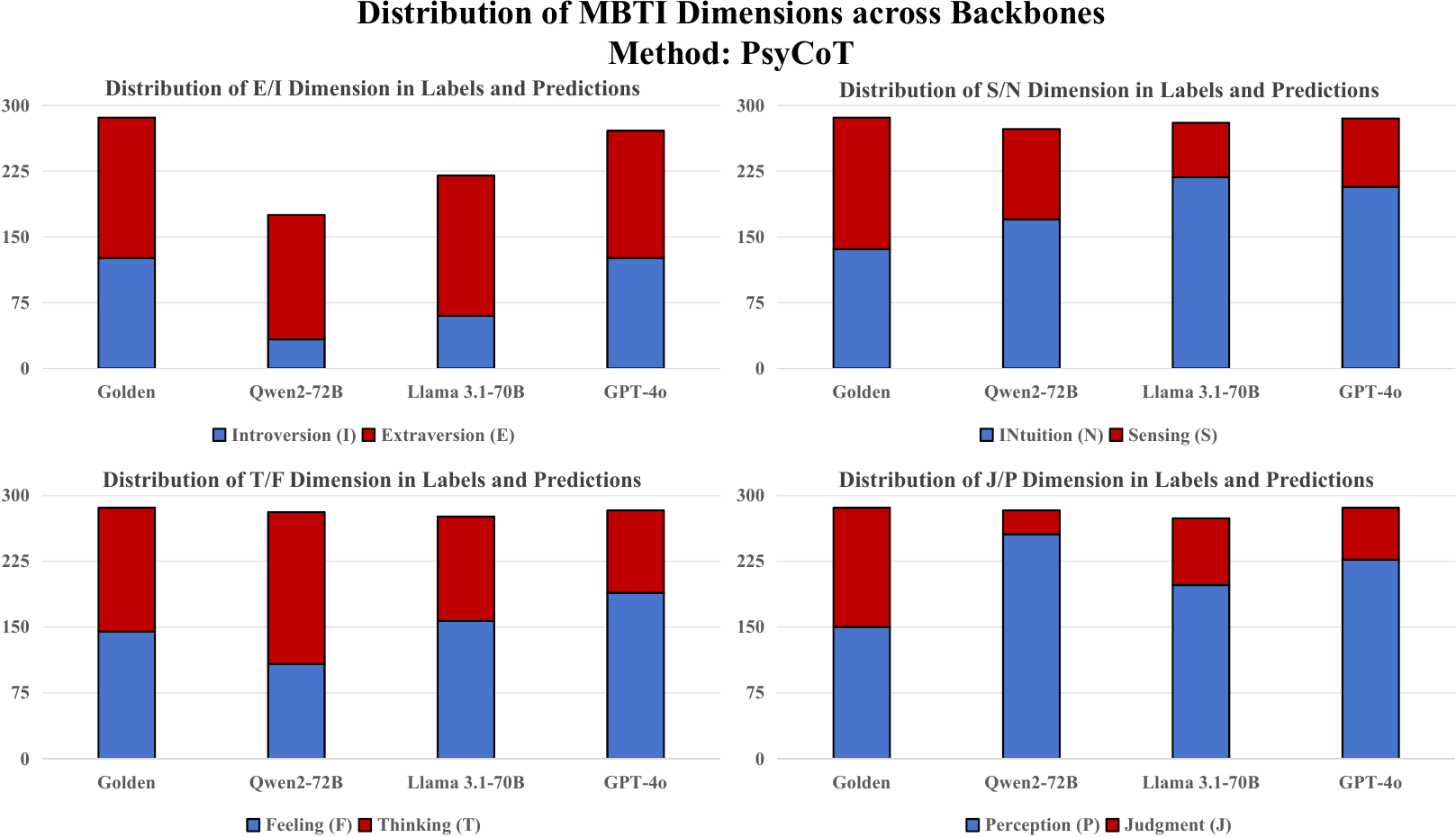}
    \caption{
Model prediction bias on \textit{PsyCoT}.
    }
\label{fig:mpsy}
\end{figure*}

We observe a significant discrepancy between Acc and F1 scores on the JP dimension across several backbones. We consider this is due to the inherent characteristics of the JP personality dimension, which introduces bias in the model's predictions. As shown in Figure \ref{fig:mstand}-\ref{fig:mpsy}, all three backbones tend to favor the P-type personality.

\begin{figure*}[!ht]
	\centering
    \vspace{0.4cm}
	\includegraphics[width=1.0\linewidth]{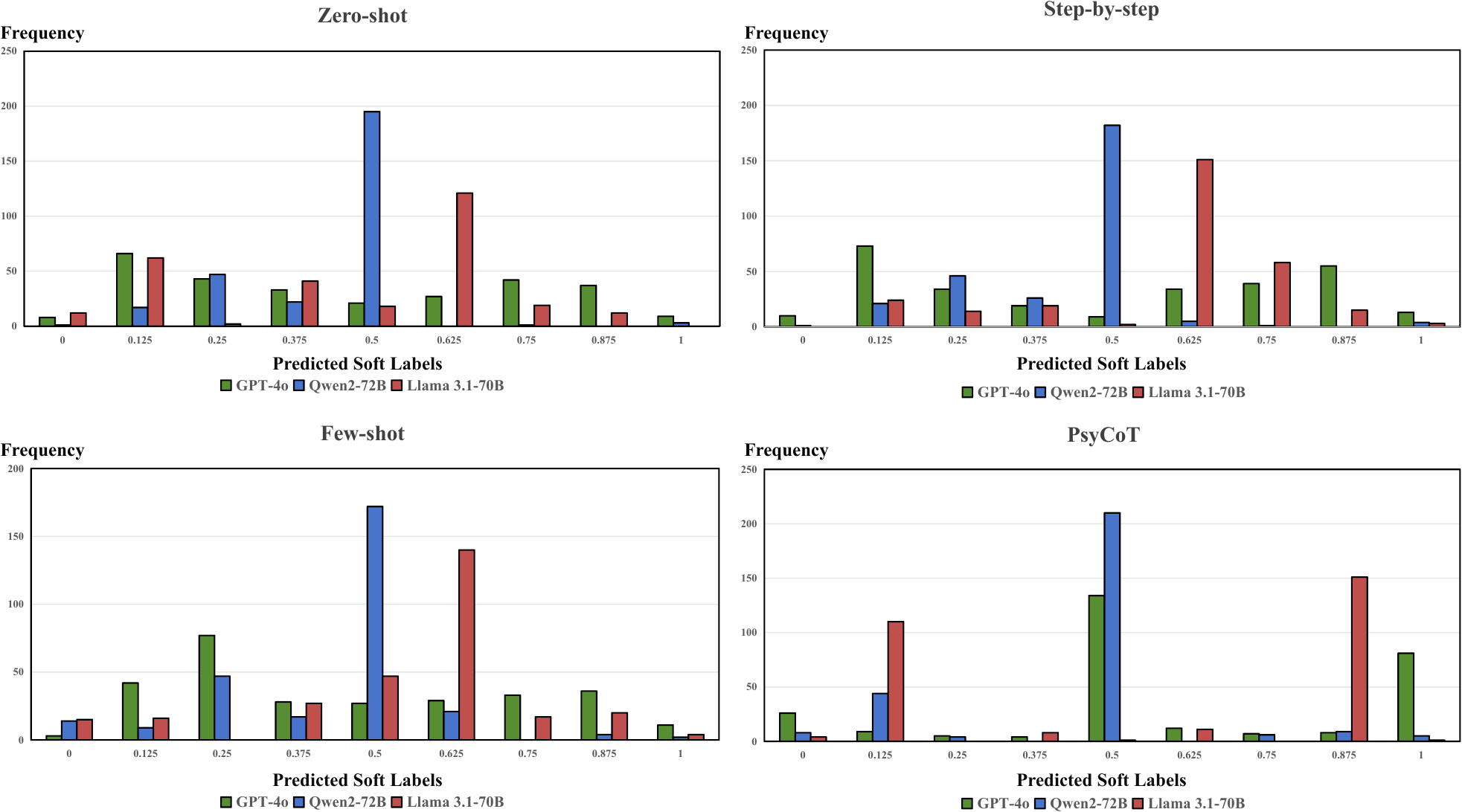}
    \caption{The score distribution of LLMs on \ourdataset for the \textit{E/I} dimension.}
    \vspace{-0.4cm}
\label{fig:finalscoreei}
\end{figure*}

\begin{figure*}[!ht]
	\centering
    \vspace{0.4cm}
	\includegraphics[width=1.0\linewidth]{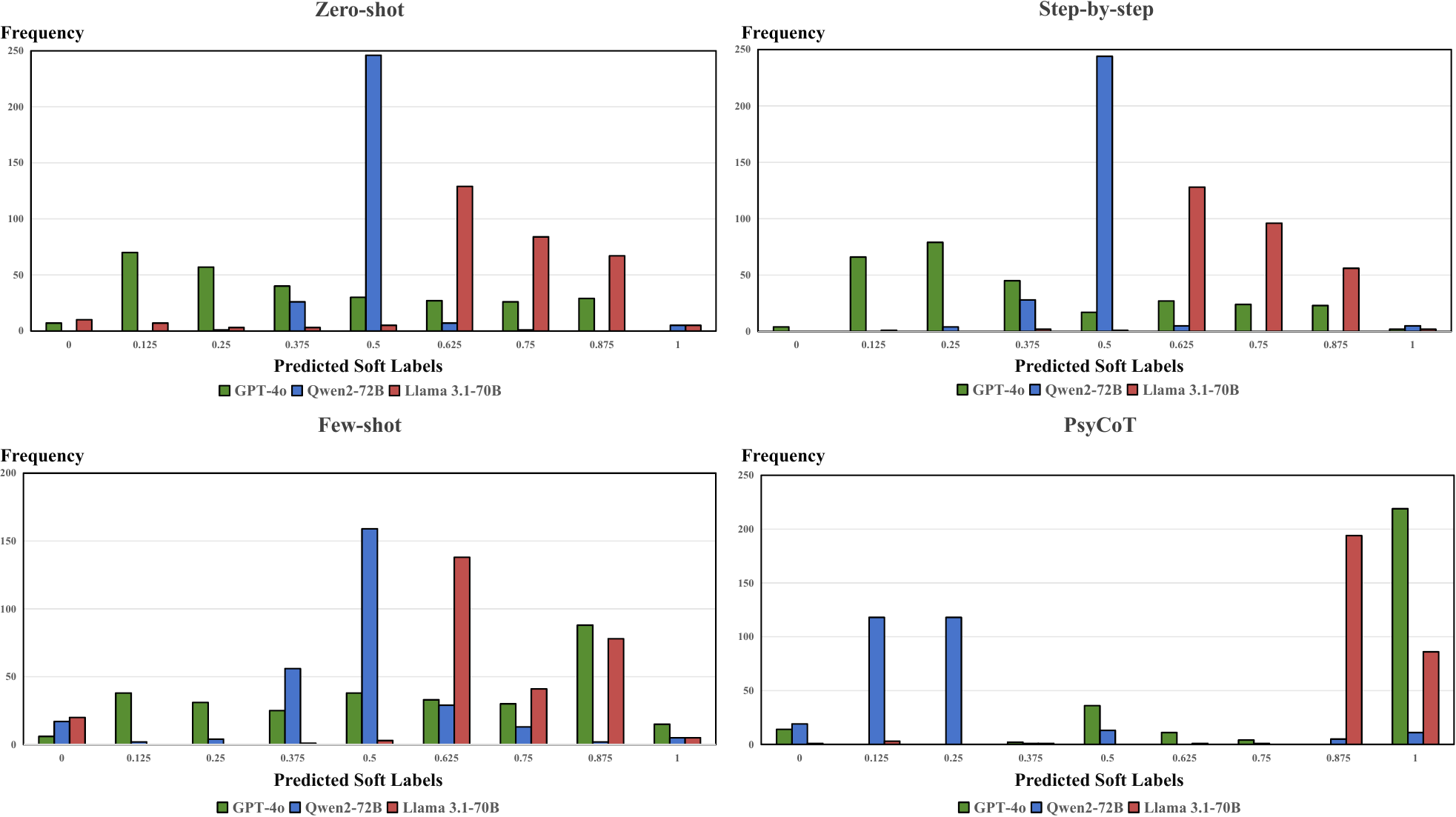}
    \caption{The score distribution of LLMs on \ourdataset for the \textit{S/N} dimension.}
    \vspace{-0.4cm}
\label{fig:finalscoresn}
\end{figure*}
\begin{figure*}[!ht]
	\centering
    \vspace{0.4cm}
	\includegraphics[width=1.0\linewidth]{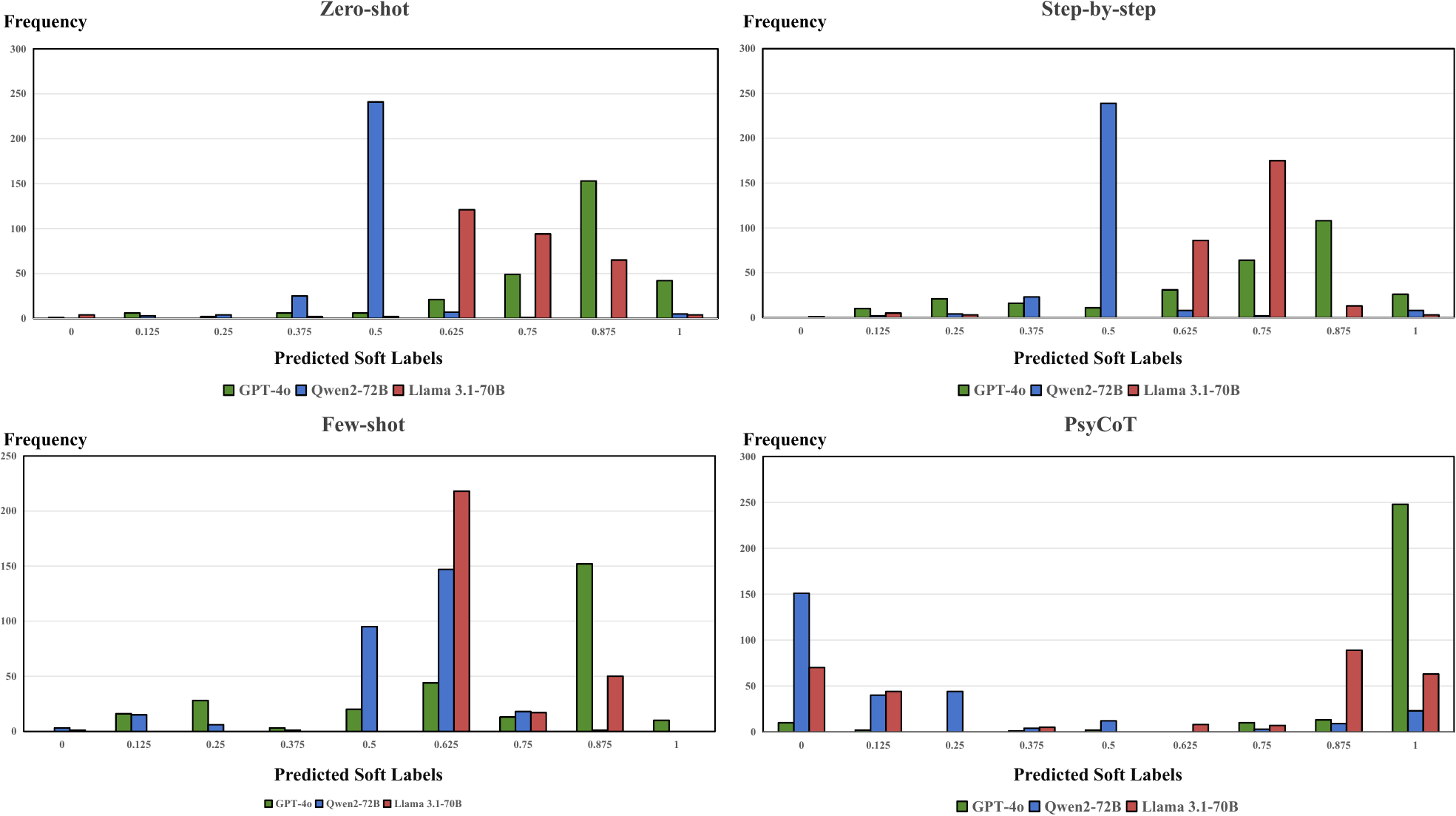}
    \caption{The score distribution of LLMs on \ourdataset for the \textit{J/P} dimension.}
    \vspace{-0.4cm}
\label{fig:finalscorejp}
\end{figure*}

\end{document}